\documentclass[10pt,journal,compsoc]{IEEEtran}
\ifCLASSOPTIONcompsoc
  \usepackage[nocompress]{cite}
\else
  \usepackage{cite}
\fi
\ifCLASSINFOpdf
   \usepackage[pdftex]{graphicx}
\else
\fi
\usepackage{amsthm,amsmath,amssymb}
\usepackage{mathrsfs}
\usepackage{booktabs}
\usepackage{multirow}
\usepackage{multicol}
\usepackage{xcolor}
\usepackage{color}
\usepackage{hyperref}

\newcommand{\tabincell}[2]{\begin{tabular}{@{}#1@{}}#2\end{tabular}}

\begin{document}
%
% paper title
% Titles are generally capitalized except for words such as a, an, and, as,
% at, but, by, for, in, nor, of, on, or, the, to and up, which are usually
% not capitalized unless they are the first or last word of the title.
% Linebreaks \\ can be used within to get better formatting as desired.
% Do not put math or special symbols in the title.
\title{SVFAP: Self-supervised Video Facial Affect Perceiver}
%
%
% author names and IEEE memberships
% note positions of commas and nonbreaking spaces ( ~ ) LaTeX will not break
% a structure at a ~ so this keeps an author's name from being broken across
% two lines.
% use \thanks{} to gain access to the first footnote area
% a separate \thanks must be used for each paragraph as LaTeX2e's \thanks
% was not built to handle multiple paragraphs
%
%
%\IEEEcompsocitemizethanks is a special \thanks that produces the bulleted
% lists the Computer Society journals use for "first footnote" author
% affiliations. Use \IEEEcompsocthanksitem which works much like \item
% for each affiliation group. When not in compsoc mode,
% \IEEEcompsocitemizethanks becomes like \thanks and
% \IEEEcompsocthanksitem becomes a line break with idention. This
% facilitates dual compilation, although admittedly the differences in the
% desired content of \author between the different types of papers makes a
% one-size-fits-all approach a daunting prospect. For instance, compsoc 
% journal papers have the author affiliations above the "Manuscript
% received ..."  text while in non-compsoc journals this is reversed. Sigh.

\author{Licai~Sun,
        Zheng~Lian,
        Kexin~Wang,
        Yu~He,
        Mingyu~Xu,
        Haiyang~Sun,
        Bin~Liu,~\IEEEmembership{Member,~IEEE,}
        and~Jianhua~Tao,~\IEEEmembership{Senior~Member,~IEEE}% <-this % stops a space
\IEEEcompsocitemizethanks{
\IEEEcompsocthanksitem Licai Sun, Kexin Wang, Yu He, Mingyu Xu, Haiyang Sun, and Biu Liu are with the School of Artificial Intelligence, University of Chinese Academy of Sciences, Beijing, China, 100049, and the National Laboratory of Pattern Recognition, Institute of Automation, Chinese Academy of Sciences, Beijing, China, 100190.\protect\\
% note need leading \protect in front of \\ to get a newline within \thanks as
% \\ is fragile and will error, could use \hfil\break instead.
E-mail: sunlicai2019@ia.ac.cn, liubin@nlpr.ia.ac.cn
\IEEEcompsocthanksitem Zheng Lian is with the National Laboratory of Pattern Recognition, Institute of Automation, Chinese Academy of Sciences, Beijing, China, 100190.\protect\\
E-mail: lianzheng2016@ia.ac.cn
% \IEEEcompsocthanksitem Zheng Lian is with National Laboratory of Pattern Recognition, Institute of Automation, Chinese Academy of Sciences, Bejing, China, 100190.\protect\\
% note need leading \protect in front of \\ to get a newline within \thanks as
% \\ is fragile and will error, could use \hfil\break instead.
% E-mail: lianzheng2016@ia.ac.cn
% \IEEEcompsocthanksitem Bin Liu is with National Laboratory of Pattern Recognition, Institute of Automation, Chinese Academy of Sciences, Bejing, China, 100190.\protect\\
% note need leading \protect in front of \\ to get a newline within \thanks as
% \\ is fragile and will error, could use \hfil\break instead.
% E-mail: liubin@nlpr.ia.ac.cn
\IEEEcompsocthanksitem Jianhua Tao is with the Department of Automation, Tsinghua University, and Beijing National Research Center for Information Science and Technology, Tsinghua University, Beijing, China, 100084.\protect\\
% note need leading \protect in front of \\ to get a newline within \thanks as
% \\ is fragile and will error, could use \hfil\break instead.
E-mail: jhtao@tsinghua.edu.cn
}% <-this % stops an unwanted space
\thanks{Manuscript received February 8, 2023, revised April 24, 2024. (Corresponding authors: Zheng Lian, Bin Liu and Jianhua Tao)}}

% note the % following the last \IEEEmembership and also \thanks - 
% these prevent an unwanted space from occurring between the last author name
% and the end of the author line. i.e., if you had this:
% 
% \author{....lastname \thanks{...} \thanks{...} }
%                     ^------------^------------^----Do not want these spaces!
%
% a space would be appended to the last name and could cause every name on that
% line to be shifted left slightly. This is one of those "LaTeX things". For
% instance, "\textbf{A} \textbf{B}" will typeset as "A B" not "AB". To get
% "AB" then you have to do: "\textbf{A}\textbf{B}"
% \thanks is no different in this regard, so shield the last } of each \thanks
% that ends a line with a % and do not let a space in before the next \thanks.
% Spaces after \IEEEmembership other than the last one are OK (and needed) as
% you are supposed to have spaces between the names. For what it is worth,
% this is a minor point as most people would not even notice if the said evil
% space somehow managed to creep in.

% The paper headers
\markboth{Journal of \LaTeX\ Class Files,~Vol.~14, No.~8, August~2015}%
{Shell \MakeLowercase{\textit{et al.}}: Bare Demo of IEEEtran.cls for Computer Society Journals}
% The only time the second header will appear is for the odd numbered pages
% after the title page when using the twoside option.
% 
% *** Note that you probably will NOT want to include the author's ***
% *** name in the headers of peer review papers.                   ***
% You can use \ifCLASSOPTIONpeerreview for conditional compilation here if
% you desire.

% The publisher's ID mark at the bottom of the page is less important with
% Computer Society journal papers as those publications place the marks
% outside of the main text columns and, therefore, unlike regular IEEE
% journals, the available text space is not reduced by their presence.
% If you want to put a publisher's ID mark on the page you can do it like
% this:
%\IEEEpubid{0000--0000/00\$00.00~\copyright~2015 IEEE}
% or like this to get the Computer Society new two part style.
%\IEEEpubid{\makebox[\columnwidth]{\hfill 0000--0000/00/\$00.00~\copyright~2015 IEEE}%
%\hspace{\columnsep}\makebox[\columnwidth]{Published by the IEEE Computer Society\hfill}}
% Remember, if you use this you must call \IEEEpubidadjcol in the second
% column for its text to clear the IEEEpubid mark (Computer Society jorunal
% papers don't need this extra clearance.)

% use for special paper notices
%\IEEEspecialpapernotice{(Invited Paper)}

% for Computer Society papers, we must declare the abstract and index terms
% PRIOR to the title within the \IEEEtitleabstractindextext IEEEtran
% command as these need to go into the title area created by \maketitle.
% As a general rule, do not put math, special symbols or citations
% in the abstract or keywords.
\IEEEtitleabstractindextext{%
\begin{abstract}
% 200 words
Video-based facial affect analysis has recently attracted increasing attention owing to its critical role in human-computer interaction. Previous studies mainly focus on developing various deep learning architectures and training them in a fully supervised manner. Although significant progress has been achieved by these supervised methods, the longstanding lack of large-scale high-quality labeled data severely hinders their further improvements. Motivated by the recent success of self-supervised learning in computer vision, this paper introduces a self-supervised approach, termed Self-supervised Video Facial Affect Perceiver (SVFAP), to address the dilemma faced by supervised methods. Specifically, SVFAP leverages masked facial video autoencoding to perform self-supervised pre-training on massive unlabeled facial videos. Considering that large spatiotemporal redundancy exists in facial videos, we propose a novel temporal pyramid and spatial bottleneck Transformer as the encoder of SVFAP, which not only largely reduces computational costs but also achieves excellent performance. To verify the effectiveness of our method, we conduct experiments on nine datasets spanning three downstream tasks, including dynamic facial expression recognition, dimensional emotion recognition, and personality recognition. Comprehensive results demonstrate that SVFAP can learn powerful affect-related representations via large-scale self-supervised pre-training and it significantly outperforms previous state-of-the-art methods on all datasets. Code is available at \textcolor[rgb]{0.93,0.0,0.47}{\url{https://github.com/sunlicai/SVFAP}}.

\end{abstract}

% Note that keywords are not normally used for peerreview papers.
}

% make the title area
\maketitle

% To allow for easy dual compilation without having to reenter the
% abstract/keywords data, the \IEEEtitleabstractindextext text will
% not be used in maketitle, but will appear (i.e., to be "transported")
% here as \IEEEdisplaynontitleabstractindextext when the compsoc 
% or transmag modes are not selected <OR> if conference mode is selected 
% - because all conference papers position the abstract like regular
% papers do.
\IEEEdisplaynontitleabstractindextext
% \IEEEdisplaynontitleabstractindextext has no effect when using
% compsoc or transmag under a non-conference mode.

% For peer review papers, you can put extra information on the cover
% page as needed:
% \ifCLASSOPTIONpeerreview
% \begin{center} \bfseries EDICS Category: 3-BBND \end{center}
% \fi
%
% For peerreview papers, this IEEEtran command inserts a page break and
% creates the second title. It will be ignored for other modes.
\IEEEpeerreviewmaketitle

\IEEEraisesectionheading{\section{Introduction}\label{sec:introduction}}
% Computer Society journal (but not conference!) papers do something unusual
% with the very first section heading (almost always called "Introduction").
% They place it ABOVE the main text! IEEEtran.cls does not automatically do
% this for you, but you can achieve this effect with the provided
% \IEEEraisesectionheading{} command. Note the need to keep any \label that
% is to refer to the section immediately after \section in the above as
% \IEEEraisesectionheading puts \section within a raised box.

% The very first letter is a 2 line initial drop letter followed
% by the rest of the first word in caps (small caps for compsoc).
% 
% form to use if the first word consists of a single letter:
% \IEEEPARstart{A}{demo} file is ....
% 
% form to use if you need the single drop letter followed by
% normal text (unknown if ever used by the IEEE):
% \IEEEPARstart{A}{}demo file is ....
% 
% Some journals put the first two words in caps:
% \IEEEPARstart{T}{his demo} file is ....
% 
% Here we have the typical use of a "T" for an initial drop letter
% and "HIS" in caps to complete the first word.

\IEEEPARstart{V}{ideo-based} facial affect analysis, which aims to automatically detect and understand human affective states from facial videos, has recently gained considerable attention due to its great potential in developing natural and harmonious human-computer interaction systems \cite{pantic2000automatic,rouast2019deep, sariyanidi2014automatic, zhao2021affective}. 
Early attempts for this task focus on designing advanced handcrafted features and machine learning algorithms on small lab-controlled datasets. With the advent of deep learning and larger labeled datasets, the research paradigm has changed to train supervised deep neural networks in an end-to-end manner. 
Researchers have developed a variety of deep architectures to improve model performance, including convolutional neural networks (CNN) \cite{fan2016video, jiang2020dfew}, recurrent neural networks (RNN) \cite{ebrahimi2015recurrent,chao2015long}, Transformers \cite{zhao2022spatial, liu2021expression}, and their combinations \cite{kollias2020exploiting, jiang2020dfew, sun2020multi, zhao2021former, ma2022spatio}.

Although tremendous progress has been achieved by supervised learning, there are still two major obstacles that impede its further development: 1) Supervised learning methods are prone to overfitting due to limited training data and the existence of label noise in current datasets \cite{li2020deep, zhang2021understanding}, thus having poor generalization ability on the unseen test set. 2) Collecting large-scale and high-quality labeled data is extremely time-consuming and labor-intensive because of the sparsity and imbalanced distribution of emotional moments in videos  \cite{li2020deep}, and also the subjectivity and ambiguity in human emotion perception \cite{lotfian2017formulating, sethu2019ambiguous}. These two irreconcilable factors make video-based facial affect analysis still far from real-world applications.

As an alternative to supervised learning, self-supervised learning has drawn massive attention recently due to its strong generalization ability and data efficiency \cite{liu2021self}. Specifically, in contrast to supervised learning, self-supervised learning leverages input data itself as the supervisory signal for self-supervised pre-training and thus can learn powerful representations from large-scale data without using any human-annotated labels \cite{jing2020self, liu2021self}. Self-supervised learning, especially generative self-supervised learning, has shown unprecedented success in lots of deep learning fields \cite{zhang2022survey}, including natural language processing and computer vision. For instance, BERT \cite{devlin2018bert} introduces masked language modeling as the pre-training objective for language representation learning and achieves state-of-the-art results on over ten natural language processing tasks. Similarly, MAE \cite{he2022masked} utilizes masked image autoencoding to perform self-supervised visual pre-training and outperforms its supervised counterpart in many downstream vision tasks.

Despite its great success in many deep learning fields, self-supervised learning has rarely been explored in video-based affective computing. 
To this end, this paper presents a self-supervised learning method, named Self-supervised Video Facial Affect Perceiver (SVFAP), to unleash the power of large-scale self-supervised learning for video-based facial affect analysis.
As shown in Fig. \ref{fig_general}, SVFAP involves two-stage training, i.e., self-supervised pre-training and downstream fine-tuning. Its whole pipeline is conceptually simple and generally inherits those of MAE \cite{he2022masked} and its video versions \cite{tong2022videomae, feichtenhofer2022masked} considering their big success in computer vision. Concretely, during the pre-training stage, SVFAP utilizes masked facial video autoencoding to learn useful and transferable spatiotemporal representations from a large amount of unlabeled facial video data. The architecture adopts an asymmetric encoder-decoder design \cite{he2022masked} to enable efficient pre-training, in which a high-capacity encoder  only processes limited visible input (since the masking ratio is very high) and a lightweight decoder operates all input and reconstructs the masked part. During fine-tuning, it discards the decoder and fine-tunes the pre-trained high-capacity encoder on downstream datasets. 
Note that, since the vanilla Vision Transformer (ViT) \cite{dosovitskiy2020image} is typically employed as the encoder in MAE and its video versions, the computational costs are very expensive during downstream fine-tuning (especially for videos) despite the architecture efficiency in pre-training. 
Considering that large redundancy (e.g., facial symmetry and temporal correlation) exists in 3D facial video data, we thus propose a novel Temporal Pyramid and Spatial Bottleneck Transformer (TPSBT) as the high-capacity encoder of SVFAP to achieve both efficient pre-training and fine-tuning. As the name suggests, TPSBT (Fig. \ref{fig_encoder}) utilizes spatial bottleneck mechanism and temporal pyramid learning to minimize redundant information in spatial and temporal dimensions, which not only reduces the computational costs greatly (about 43\% FLOPs reduction during fine-tuning) but also leads to superior performance.
% 

% main contributions
To verify the effectiveness of SVFAP, we conduct experiments on nine datasets from three video-based facial affect analysis tasks, including six datasets for dynamic facial expression recognition, two datasets for dimensional emotion recognition, and one dataset for personality recognition. Comprehensive experimental results demonstrate that our SVFAP can learn powerful affect-related representations from large-scale unlabeled facial video data via self-supervised pre-training and significantly outperforms previous state-of-the-art methods on all downstream datasets. 
For instance, on three in-the-wild dynamic facial expression recognition datasets, our best model surpasses the previous best by 5.72\% UAR and 5.02\% WAR on DFEW \cite{jiang2020dfew}, 4.38\% UAR and 3.75\% WAR on FERV39k \cite{wang2022ferv39k}, and 7.91\% UAR and 6.10\% WAR on MAFW \cite{liu2022mafw}. 
To sum up, the main contributions of this paper are three-fold:
\begin{itemize}
\item We introduce a self-supervised learning approach, SVFAP, to address the dilemma faced by supervised learning methods in video-based facial affect analysis. 
It leverages masked facial video autoencoding as the pre-training objective and can learn powerful affect-related representations from large-scale unlabeled facial video data. 
\item We propose a novel TPSBT model as the encoder of SVFAP to eliminate large spatiotemporal redundancy in facial videos, which not only enjoys lower computational costs but also has superior performance when compared with the vanilla ViT.
\item Comprehensive experiments on nine downstream datasets demonstrate that our SVFAP achieves state-of-the-art performance in three popular  video-based facial affect analysis tasks.
\end{itemize}

\section{Related Work}

\subsection{Video-based Facial Affect Analysis}
Most studies on video-based facial affect analysis fall into the supervised learning paradigm. They mainly concentrate on developing more advanced deep learning architectures to extract discriminative spatiotemporal representations from raw facial videos. 
Generally, there are two lines of research. 
The first line of research treats the spatial and temporal dimension of 3D video data in an independent manner \cite{kollias2020exploiting, jiang2020dfew, sun2020multi, liu2022mafw, wang2022ferv39k, wang2022dpcnet}. Typically, 2D CNN (e.g., VGGNet \cite{simonyan2014very} and ResNet \cite{he2016deep}) is first used to extract spatial features from each static frame and then RNN (e.g., LSTM \cite{hochreiter1997long} and GRU \cite{chung2014empirical}) runs over them to integrate the dynamic temporal information across all frames. Recently, inspired by the great success of Transformer \cite{vaswani2017attention} in natural language processing and computer vision, there are also several studies that utilize the global dependency modeling ability of Transformer to enhance spatial and temporal feature extraction of traditional CNN and RNN \cite{zhao2021former, liu2021expression, ma2022spatio, zhao2022spatial, li2023intensity}. 
Another line of research tries to simultaneously encode spatial appearance features and dynamic motion information by extending the 2D convolution
kernel to the 3D convolution kernel along the temporal axis. With the help of 3D kernels, 3D CNN (e.g., C3D \cite{tran2015learning}, R(2+1)D \cite{carreira2017quo}, P3D \cite{qiu2017learning}, and 3D ResNet \cite{hara2018can}) is expected to extract discriminative spatiotemporal representations from raw videos. 

Although the above supervised deep learning methods have achieved remarkable improvement over traditional machine learning methods, they still suffer from the notorious overfitting issue due to limited training data and label noise in current datasets \cite{li2020deep, zhang2021understanding}. 
In contrast to these supervised methods, we propose a self-supervised learning method in this study to address their dilemma by pre-training on abundantly available unlabeled facial videos. 

% \begin{figure*}[t]
% 	\centering
% 	\includegraphics[width=1.0\linewidth]{imgs/general.pdf}
% 	\caption{An overview of the proposed method (i.e, SVFAP). It consists of two stages, including self-supervised pre-training (a) and downstream fine-tuning (b). During pre-training, SVFAP utilizes masked facial video autoencoding as the training objective. Following previous studies \cite{he2022masked, tong2022videomae, feichtenhofer2022masked}, it adopts an asymmetric encoder-decoder architecture and a high masking ratio (e.g., 90\%) to enable fast pre-training on large-scale unlabeled facial video data. After pre-training, the lightweight decoder is discarded and the high-capacity encoder is fine-tuned in downstream tasks.}
% 	\label{fig_general}
% \end{figure*}

\subsection{Self-supervised Learning}
Self-supervised learning aims to solve the data-hungry issue in supervised learning by exploiting massive unlabeled data.
It can be roughly divided into two categories: discriminative and generative \cite{zhang2022survey}. 
The discriminative method generally follows the supervised counterpart by designing a discriminative loss. Early studies focused on developing various geometry-based pretext tasks, such as predicting image rotation \cite{komodakis2018unsupervised} and sorting shuffled
video frames \cite{lee2017unsupervised}. In recent years, the trend has shifted from handcrafted tasks to contrastive learning methods, e.g., MoCo \cite{he2020momentum}, and SwAV \cite{caron2020unsupervised}. 
Contrastive learning has been the dominant self-supervised pre-training framework in computer vision until the more recent reviving success of generative learning. 
The generative method typically involves an autoencoding process, in which an encoder maps the input into a latent representation and a decoder reconstructs the original input from the encoded representation. 
Early generative work can be dated back to denoising autoencoder \cite{vincent2008extracting}, i.e., reconstructing the clean input from a partially corrupted one. 
Recently, motivated by the unprecedented success in natural language processing (e.g., BERT \cite{devlin2018bert} and GPT \cite{radford2018improving}), generative methods have also achieved impressive results in computer vision \cite{bao2021beit, he2022masked}. 
One of the most representative methods is masked autoencoder (MAE) \cite{he2022masked}. MAE and its video versions \cite{tong2022videomae, feichtenhofer2022masked} utilize an asymmetric encoder-decoder architecture to efficiently pre-train vanilla ViT and outperform its supervised counterpart and state-of-the-art contrastive learning methods by large margins in many downstream tasks (e.g., object recognition/detection/segmentation, and action recognition/detection).

Despite the great success of self-supervised learning in many deep learning fields, it has rarely been explored in video-based facial affect analysis. Roy et al. \cite{roy2021spatiotemporal} present a spatiotemporal contrastive learning method pre-trained on a small lab-controlled labeled dataset. 
Unlike it, the proposed SVFAP is built upon more advanced generative methods (i.e., MAE and its video versions) and leverages large-scale in-the-wild unlabeled facial videos for self-supervised pre-training.
There are also several self-supervised studies in the relevant research field (i.e., facial action unit detection). For instance,  
FAb-Net \cite{wiles2018self} exploits relative facial movements between adjacent frames as free supervisory signals to perform self-supervised pre-training. TCAE \cite{li2019self} improves FAb-Net by disentangling the head pose-related movements and facial action-related ones. FaceCycle \cite{chang2021learning} further introduces facial motion and identity cycle-consistency to promote facial representation learning. Lu et al. \cite{lu2020self} propose a triplet-based frame ranking method for temporal consistency modeling. 
Recently, there are also a few studies focusing on general self-supervised facial representation learning.
Bulat et al. \cite{bulat2022pre} pre-train SwAV \cite{caron2020unsupervised} on large-scale face recognition datasets and find that it achieves significant improvements over previous supervised methods on five face analysis tasks.
FaRL \cite{zheng2022general} utilizes both low-level masked facial image modeling and high-level face-text contrastive learning for self-supervised pre-training and outperforms state-of-the-art methods on many face analysis tasks. 
Although achieving promising results, these methods use 2D models for self-supervised pre-training, thus could not capture rich spatiotemporal information in 3D facial videos.

\subsection{Vision Transformer}
Originated from natural language processing, Transformer \cite{vaswani2017attention} based architectures have recently revolutionized various computer vision tasks by means of its strong long-range dependency modeling ability \cite{han2022survey}.
Among them, the pioneering work of ViT \cite{dosovitskiy2020image} directly applies the standard Transformer to a sequence of image patches and performs very well on image classification tasks, challenging the dominant paradigm of CNN in computer vision. Since ViT relies on large amounts of labeled data (i.e., JFT-300M) to achieve successful supervised pre-training, DeiT \cite{touvron2021training} introduces several training strategies to allow it can be trained on a much smaller dataset (i.e., ImageNet-1K). 
After that, numerous ViT variants have emerged by incorporating more or less vision-friendly priors (e.g., local self-attention and hierarchical design) to reduce the quadratic scaling cost of vanilla ViT and improve model performance in vision tasks \cite{liu2021swin, bertasius2021space, fan2021multiscale, li2022mvitv2, liu2022video}. For instance, in the video domain, 
TimeSformer \cite{bertasius2021space} applies factorized temporal and spatial attention in the standard Transformer block to achieve spatiotemporal representation learning. 
MViT \cite{fan2021multiscale} introduces multi-scale feature hierarchies to the vanilla ViT.
Video Swin Transformer\cite{liu2022video} utilizes 3D-shifted window attention to inject spatiotemporal locality inductive bias into video Transformers. Although these variants usually perform better than vanilla ViT in the supervised setting, they are typically not suitable for masked autoencoding in self-supervised pre-training.
This is because they could not drop masked tokens due to the incorporation of local operations (e.g., window-based attention and patch merging) in their architectures. Thus, they can not enjoy the efficiency of vanilla ViT during masked autoencoding. 
This is why MAE and its video versions all employ ViT as the encoder. Nevertheless, ViT still suffers from expensive computational costs during downstream fine-tuning. 
To this end, in this paper, we introduce temporal pyramid learning and spatial bottleneck mechanism to ViT to enable both efficient pre-training and fine-tuning.

\section{Method}

\begin{figure*}[t]
	\centering
	\includegraphics[width=1.0\linewidth]{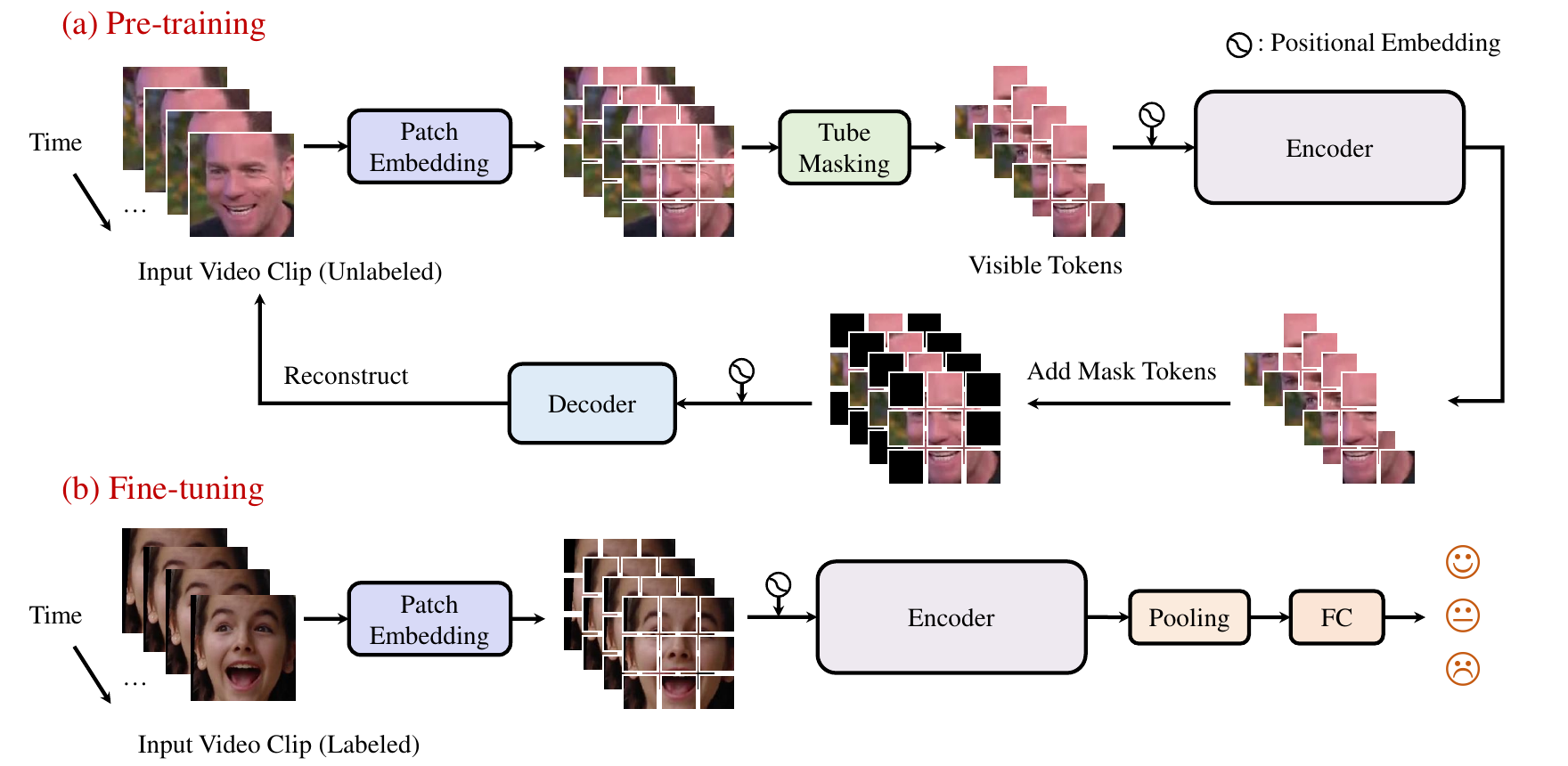}
	\caption{An overview of the proposed method (i.e., SVFAP). It consists of two stages, including self-supervised pre-training (a) and downstream fine-tuning (b). During pre-training, SVFAP utilizes masked facial video autoencoding as the training objective. Following previous studies \cite{he2022masked, tong2022videomae, feichtenhofer2022masked}, it adopts an asymmetric encoder-decoder architecture and a high masking ratio (e.g., 90\%) to enable fast pre-training on large-scale unlabeled facial video data. After pre-training, the lightweight decoder is discarded and the high-capacity encoder is fine-tuned in downstream tasks.}
	\label{fig_general}
\end{figure*}

\begin{figure*}[t]
	\centering
	\includegraphics[width=1.0\linewidth]{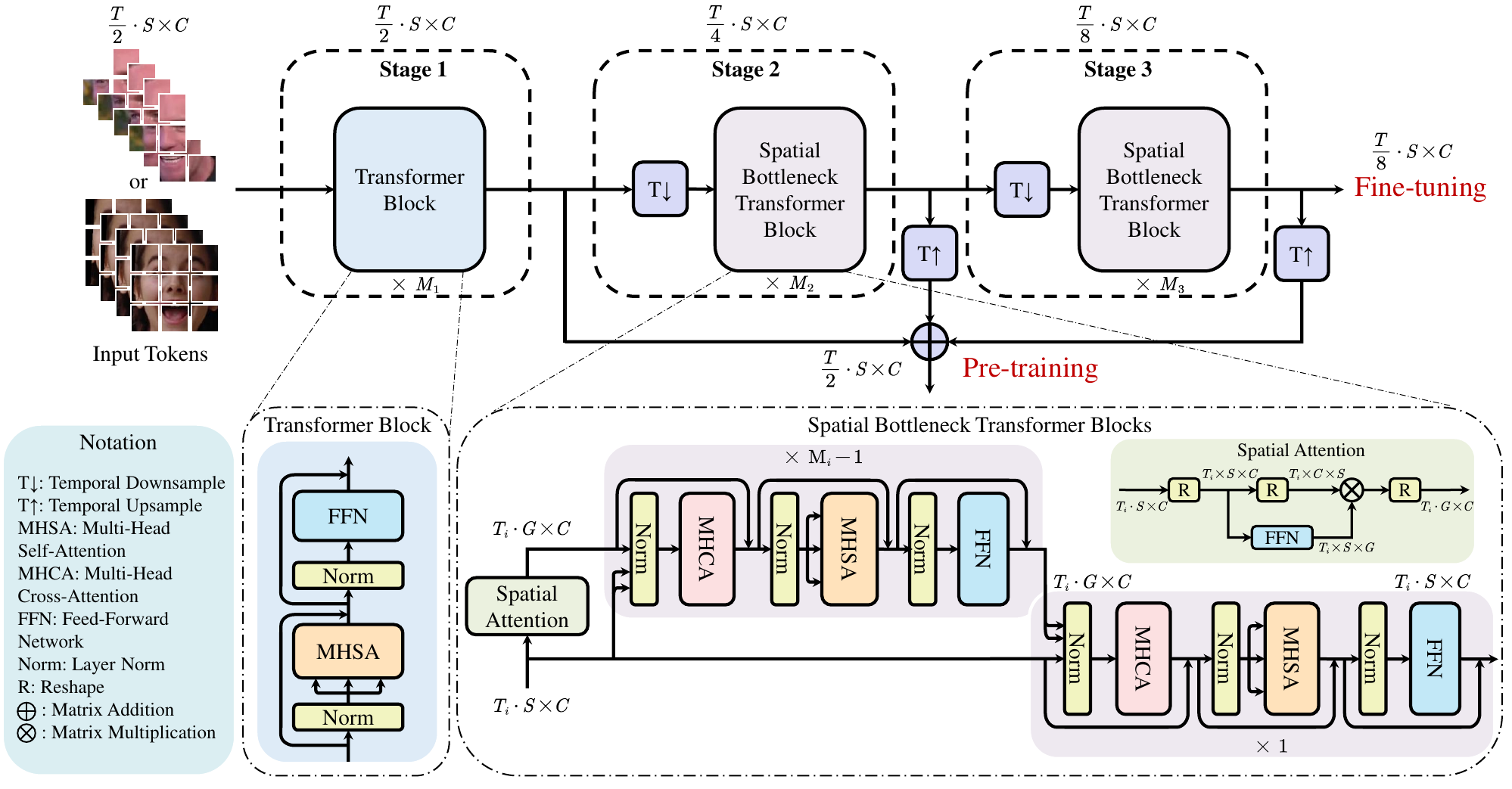}
	\caption{The encoder in SVFAP, i.e., TPSBT. We perform summation-based multi-scale fusion for features from three stages during pre-training and empirically do not use it in fine-tuning. The temporal length in each stage $T_i = \frac{T}{2k^{i-1}}$ ($k$ is the downsampling rate and empirically $k=2$, $i \in \{1,2,3\}$). The spatial size $S = \frac{H}{16} \cdot \frac{W}{16} \cdot (1-\rho)$ ($\rho$ is the masking ratio and $\rho=90\%$) during pre-training and $S=\frac{H}{16} \cdot \frac{W}{16}$ during fine-tuning. }
	\label{fig_encoder}
\end{figure*}

To tackle the longstanding data scarcity issue in video-based affective computing, we propose Self-supervised Video Facial Affect Perceiver (SVFAP) in this paper, with its whole pipeline illustrated in Fig. \ref{fig_general}. 
Specifically, it first utilizes masked facial video autoencoding to perform self-supervised pre-training on massive unlabeled facial video data, then discards the decoder and fine-tunes the pre-trained encoder in downstream video-based facial affect analysis tasks. In the following sections, we elaborate on both pre-training and fine-tuning details.

\subsection{Self-supervised Pre-training}
The idea of masked facial video autoencoding is simple, i.e., reconstructing the original facial video given partially observed input. Concretely, SVFAP consists of an encoder to project the original video to a high-level spatiotemporal representation in the latent space, and then a decoder to reconstruct the original input from the encoded representation. Moreover, similar to MAE \cite{he2022masked} and its video versions \cite{tong2022videomae, feichtenhofer2022masked}, SVFAP adopts an asymmetric encoder-decoder architecture, where a high-capacity encoder only accepts limited visible tokens (e.g., 10\%) as input and a lightweight decoder operates on all tokens to perform the video reconstruction. With this special design, the computational costs could be largely reduced to allow fast pre-training.

\subsubsection{Patch Embedding}
We employ a Transformer-based encoder to encode raw facial videos. Considering that Transformer accepts a sequence of tokens as input, we thus split the 3D video clip $\mathbf{V}$ with a spacetime shape of $T \times H \times W \times 3$ ($T$ is the number of frames, $H$ and $W$ are frame height and width respectively) into a regular grid of non-overlapping spatiotemporal patches (Fig. \ref{fig_general} (a)), then flatten and embed them to tokens by linear projection. Following VideoMAE \cite{tong2022videomae}, the patch size is set to $2 \times 16 \times 16$. Therefore, the raw video clip is transformed to token embeddings with the shape of $\frac{T}{2} \times \frac{H}{16} \times \frac{W}{16} \times C$ ($C$ is the embedding dimension) after patch embedding. Finally, to inject the position information into the sequence, we add sinusoidal positional embeddings \cite{vaswani2017attention} to the embedded tokens.

\subsubsection{Tube Masking}
In order to make video reconstruction a challenging and meaningful pre-training task, we need to mask some embedded patches. Different modalities usually require different masking ratios. For the high-level and information-dense text modality, BERT \cite{devlin2018bert} chooses to randomly mask 15\% of tokens in the sequence. For the low-level and information-redundant image modality, MAE advocates a masking ratio of 75\%. 
For video data, it has one more temporal dimension compared with static images and has high spatiotemporal redundancy.
The empirical findings in previous studies suggest that 90\% works best for video data \cite{tong2022videomae, feichtenhofer2022masked}. Following them, we also adopt a masking ratio of $\rho=90\%$. With such a high masking ratio, most masked tokens can be dropped and only a small subset (i.e., 10\%) of visible tokens will be processed by the encoder, leading to large computational cost reduction and efficient pre-training. 

Moreover, considering that consecutive frames in videos are highly similar, spacetime-agnostic random masking might incur information leakage and make masked video reconstruction too easy. The reason is that the model could easily reconstruct the masked patches by simply finding similar visible ones in adjacent frames. Thus, the shortcut signal will encourage the model to capture low-level temporal correspondence and undermine the desired high-level spatiotemporal structure learning. To address this issue, we follow VideoMAE to employ a simple masking strategy, termed tube masking or space-only masking, i.e., first masking spatial patches in one temporal slice and then sharing it along the temporal axis (Fig. \ref{fig_general} (a)).

\subsubsection{Encoder}
Typically, the vanilla ViT \cite{dosovitskiy2020image} is employed as the encoder in masked autoencoding \cite{he2022masked, tong2022videomae, feichtenhofer2022masked}. Although it enjoys much efficiency during pre-training by discarding many masked tokens, its computational costs are still very expensive during downstream fine-tuning as it must take as input all embedded tokens. Meanwhile, we notice that large redundancy exists in 3D facial videos, such as facial symmetry and temporal correlation.
Motivated by this, we propose the Temporal Pyramid and Spatial Bottleneck Transformer (TPSBT) as the encoder of our SVFAP to enable both fast pre-training and fine-tuning. 
As shown in Fig. \ref{fig_encoder}, our TPSBT has three stages. The first stage is composed of several standard Transformer blocks, while in the last two stages, they are replaced by spatial bottleneck Transformer blocks to reduce spatial redundancy. Moreover, benefiting from tube masking, we could further utilize temporal pyramid learning to reduce redundancy in the temporal dimension. For convenience, in the following part, we denote as $\mathbf{X} \in \mathbb{R}^{N \times C}$ the reshaped input token embeddings after tube masking, where $N = \frac{T}{2} \cdot S$ is the total number of visible tokens and $S = \frac{H}{16} \cdot \frac{W}{16} \cdot (1-\rho$) denotes the spatial token number. And $\mathbf{X}_i \in \mathbb{R}^{N_i \times C}$ represents the output of stage $i$ ($N_i$ is the number of tokens in each stage, $i \in \{1, 2, 3\}$).   

\textbf{Standard Transformer}. 
The standard Transformer in stage 1 is used to retain the global spatiotemporal representation learning ability of vanilla ViT. It consists of a sequence of $M_1$ Transformer blocks, each of which is composed of alternating layers of Multi-Head Self-Attention (MHSA) and Feed-Forward Network (FFN), with Layer Normalization (LN) applied before each layer and residual connections after each layer. 
Formally, we define a standard Transformer block as follows:
\begin{equation}
\begin{split}
\mathbf{Y}^{(j-1)}_1 &= \textrm{MHSA}(\textrm{LN}(\mathbf{X}^{(j-1)}_1)) + \mathbf{X}^{(j-1)}_1 \\
\mathbf{X}^{(j)}_1 &= \textrm{FFN}(\textrm{LN}(\mathbf{Y}^{(j-1)}_1)) + \mathbf{Y}^{(j-1)}_1
\end{split}
\label{eq_transformer}
\end{equation}
where the superscript $j \in \{1,..., M_1\}$ is the block index, $\mathbf{X}^{(0)}_1 = \mathbf{X}$, and the output of stage 1 is $\mathbf{X}_1 = \mathbf{X}^{(M_1)}_{1}$.

The MHSA operation in Eq. (\ref{eq_transformer}) employs multi-head scaled dot-product attention \cite{vaswani2017attention} to explore long-range spatiotemporal dependencies in input tokens, i.e.,
\begin{equation}
\begin{split}
\textrm{MHSA}(\mathbf{X}) &= \textrm{Concat}(\textrm{head}_{1}, ..., \textrm{head}_{H})\mathbf{W}^O \\
\textrm{head}_{h} &= \textrm{Softmax}(\frac{\mathbf{Q}_h \mathbf{K}_h^\top}{\sqrt{d_h}})\mathbf{V}_h
\end{split}
\label{eq_mhsa}
\end{equation}
where $\mathbf{Q}_h = \mathbf{X} \mathbf{W}^{Q}_h$ is the query, $\mathbf{K}_h = \mathbf{X}  \mathbf{W}^{K}_h$ is the key, $\mathbf{V}_h = \mathbf{X} \mathbf{W}^{V}_h$ is the value, $\mathbf{W}^{*}_h \in \mathbb{R}^{C\times d_h}$ ($* \in \{Q,K,V \}$), $\mathbf{W}^{O} \in \mathbb{R}^{C\times C}$, $H$ is the number of heads, $d_h=C/H$ is the feature dimension of head $h$, and the superscript $^\top$ is matrix transpose. Finally, FFN in Eq. (\ref{eq_transformer}) consists of two linear layers with a GELU \cite{hendrycks2016gaussian} non-linearity in between, i.e., 
\begin{equation}
\textrm{FFN}(\mathbf{X}) = \textrm{GELU}(\mathbf{X} \mathbf{W}_1+\mathbf{b}_1) \mathbf{W}_2 + \mathbf{b}_2
\label{eq_ffn}
\end{equation}
where $\mathbf{W}_1 \in \mathbb{R}^{C\times 4C}$, $\mathbf{b}_1 \in \mathbb{R}^{4C}$, $\mathbf{W}_2 \in \mathbb{R}^{4C\times C}$, $\mathbf{b}_2 \in \mathbb{R}^{C}$.

\textbf{Temporal Pyramid}.
As illustrated in Fig. \ref{fig_encoder}, a temporal downsampling module is inserted before spatial bottleneck Transformer blocks in the last two stages to achieve temporal redundancy reduction. We utilize a strided convolution layer to implement this module:
\begin{equation}
\begin{split}
\textrm{T}\downarrow(\mathbf{X}_{i-1}) &= \textrm{Conv}(\mathbf{X}_{i-1}, k) \in \mathbb{R}^{N_i \times C} \\
\end{split}
\label{eq_td}
\end{equation}
where $i \in \{2,3\}$ is the stage index, $N_i= T_i \cdot S$, $T_i = \frac{T}{2k^{i-1}}$ is the temporal length in stage $i$, and $k$ is the kernel size and stride of the convolution layer. Empirically, $k$ is set to 2, which means that the temporal length is halved after the temporal downsampling module. Alternatives to convolution-based downsampling could be more simple average pooling or max pooling, however, we observed slight performance degradation in experiments. 

\textbf{Spatial Bottleneck Transformer}.
After temporal information aggregation, we further employ the Spatial Bottleneck Transformer (SBT) in stage 2 and stage 3 to reduce spatial redundancy. The main idea of SBT is to introduce a small number of bottleneck tokens via spatial attention and employ them instead of the original redundant embedded tokens to perform further spatiotemporal interactions.

As shown in the bottom right of Fig. \ref{fig_encoder}, SBT is composed of a spatial attention module, $M_i-1$ ($i \in \{2,3\}$) SBT blocks, and 1 reverse SBT block.
To be specific, the spatial attention module is used to compress the original fine-grained and redundant spatial tokens into a few global semantic bottleneck tokens. 
For convenience, we reuse $\mathbf{X}_{i-1} \in \mathbb{R}^{T_i \cdot S \times C}$ ($i \in \{2,3\}$) as the notation of the output of temporal downsampling module in Eq. (\ref{eq_td}). Then, the spatial attention is formulated as follows:
\begin{equation}
\begin{split}
\hat{\mathbf{X}}_{i-1} &= \textrm{Reshape}(\mathbf{X}_{i-1}) \in \mathbb{R}^{T_i \times S \times C} \\
\mathbf{Y}_i &= \textrm{GELU}(\hat{\mathbf{X}}_{i-1} \mathbf{W}_1 + \mathbf{b}_1) \mathbf{W}_2 + \mathbf{b}_2 \in \mathbb{R}^{T_i \times S \times G} \\
\tilde{\mathbf{X}}_{i-1} &= \textrm{Reshape}(\mathbf{X}_{i-1}) \in \mathbb{R}^{T_i \times C \times S} \\
\mathbf{Z}_i &= \Tilde{\mathbf{X}}_{i-1} \mathbf{Y}_i \in \mathbb{R}^{T_i \times C \times G} \\
\mathbf{B}_i &= \textrm{Reshape}(\mathbf{Z}_i) \in \mathbb{R}^{T_i \cdot G \times C}
\end{split}
\label{eq_sa}
\end{equation}
where $\mathbf{W}_1$, $\mathbf{b}_1$, $\mathbf{W}_2$, and $\mathbf{b}_2$ have similar feature dimensions to those in Eq. (\ref{eq_ffn}). After spatial attention, the redundant $S$ spatial tokens of each temporal slice in the original input $\mathbf{X}_{i-1}$ are summarized into $G$  bottleneck tokens in $\mathbf{B}_i$.

Subsequently, we utilize $M_i-1$ SBT blocks to operate on the summarized bottleneck tokens to achieve efficient global spatiotemporal interactions. 
As shown in Fig. \ref{fig_encoder}, an SBT block mainly consists of a Multi-Head Cross-Attention (MHCA) layer \cite{jaegle2021perceiver}, an MHSA layer, and an FFN layer. Formally, it can be defined as follows:
\begin{equation}
\begin{split}
\mathbf{Y}^{(j-1)}_i &= \textrm{MHCA}(\textrm{LN}(\mathbf{B}^{(j-1)}_i), \textrm{LN}(\mathbf{X}_{i-1})) + \mathbf{B}^{(j-1)}_i \\
\mathbf{Z}^{(j-1)}_i &= \textrm{MHSA}(\textrm{LN}(\mathbf{Y}^{(j-1)}_i)) + \mathbf{Y}^{(j-1)}_i \\
\mathbf{B}^{(j)}_i &= \textrm{FFN}(\textrm{LN}(\mathbf{Z}^{(j-1)}_i)) + \mathbf{Z}^{(j-1)}_i
\end{split}
\label{eq_sbt}
\end{equation}
where the superscript $j \in \{1, ..., M_i - 1\}$ is the block index, $\mathbf{B}^{(0)}_i = \mathbf{B}_i$.
The MHCA layer iteratively distills necessary information from the original input back into the bottleneck embeddings to avoid key information being lost during spatial compression. It is a variant of MHSA, whose query is the bottleneck embeddings while the key and value come from the original input to allow information flow from the latter to the former, i.e.,
\begin{equation}
\begin{split}
\textrm{MHCA}(\mathbf{X}, \mathbf{Y}) &= \textrm{Concat}(\textrm{head}_{1}, ..., \textrm{head}_{H})\mathbf{W}^O \\
\textrm{head}_{h} &= \textrm{Softmax}(\frac{\mathbf{Q}_h^X {\mathbf{K}_h^Y}^\top}{\sqrt{d_h}})\mathbf{V}_h^Y
\end{split}
\label{eq_mhca}
\end{equation}
where $\mathbf{Q}_h^X = \mathbf{X} \mathbf{W}^{Q}_h$, $\mathbf{K}_h^Y = \mathbf{Y}  \mathbf{W}^{K}_h$, $\mathbf{V}_h^Y = \mathbf{Y} \mathbf{W}^{V}_h$ , other notations are similar to those in Eq. (\ref{eq_mhsa}).

% complexity analysis
Compared with a standard Transformer block, the computational complexity of an SBT block is reduced from the quadratic $O(S^2T_i^2)$ to $O(G(G+S)T_i^2)$ thanks to the introduction of spatial bottleneck tokens. Considering that $G$ is a small constant, an SBT block thus enjoys linear computational complexity with respect to the spatial size $S$. Note that, empirically, we have $G \approx S$ during pre-training and $G \ll S$ during fine-tuning (Sec. \ref{sec_impl}). 
Therefore, the computational costs of pre-training remain approximately unchanged (actually, slightly decrease) and large computations can be reduced during downstream fine-tuning, allowing both efficient pre-training and fine-tuning (Sec. \ref{sec_ablation}).

Ultimately, a reverse SBT block is employed to recover the spatial resolution of the original input for the final reconstruction. 
As shown in Fig .\ref{fig_encoder}, the query and key/value of this reverse block are just the opposite of the normal block, i.e., 
\begin{equation}
\begin{split}
\mathbf{Y}_i &= \textrm{MHCA}(\textrm{LN}(\mathbf{X}_{i-1}), \textrm{LN}(\mathbf{B}^{(M_i - 1)}_i)) + \mathbf{X}_{i-1} \\
\mathbf{Z}_i &= \textrm{MHSA}(\textrm{LN}(\mathbf{Y}_i)) + \mathbf{Y}_i \\
\mathbf{X}_{i} &= \textrm{FFN}(\textrm{LN}(\mathbf{Z}_i)) + \mathbf{Z}_i
\end{split}
\label{eq_sbt}
\end{equation}
where $\mathbf{X}_{i} \in \mathbb{R}^{N_i \times C}$ is the final output of stage $i$ ($i \in \{2,3\}$).

\subsubsection{Decoder}
The decoder is only utilized during pre-training to reconstruct the original input facial video. Therefore, it is flexible to build the decoder architecture in a way that is independent of the encoder design. For simplicity, we follow previous work \cite{he2022masked, tong2022videomae} to employ the standard Transformer. 
Moreover, since it has demonstrated that the decoder can be lightweight, the number of Transformer blocks is set to 4, which is much smaller than that of the encoder (Sec. \ref{sec_impl}). 

As illustrated in Fig. \ref{fig_general} (a), the input to the decoder is the combination of visible and masked tokens. It should be noted that, due to the existence of temporal downsampling modules in the encoder, it is necessary to recover the temporal resolution of visible tokens. Hence, we apply corresponding temporal upsampling to the outputs of the last two stages. To avoid increasing model parameters, we simply use nearest-neighbor interpolation for upsampling:
\begin{equation}
\begin{split}
\textrm{T}\uparrow(\mathbf{X}_{i}) &= \textrm{Interp}(\mathbf{X}_{i}, k^{i-1}) \in \mathbb{R}^{N \times C} \\
\end{split}
\label{eq_tu}
\end{equation}
where $\mathbf{X}_i$ is the output of stage $i$ ($i \in \{2,3\}$) in the encoder, $k^{i-1}$ is the upscaling factor. 
Moreover, to enable the decoder to be aware of spatiotemporal features at different levels, summation-based multi-scale fusion is employed to aggregate the outputs of three stages in the encoder, i.e.,
\begin{equation}
\begin{split}
\mathbf{X} = \mathbf{X}_1 + \textrm{T}\uparrow(\mathbf{X}_2) + \textrm{T}\uparrow(\mathbf{X}_3) 
\end{split}
\label{eq_tu}
\end{equation}
where $\mathbf{X} \in \mathbb{R}^{N \times C}$ denotes the encoded visible tokens.
After multi-scale fusion, we first concatenate $\mathbf{X}$ and the trainable masked tokens, then add sinusoidal positional embeddings to them, and finally pass them through the Transformer-based decoder for video reconstruction. Finally, the mean squared error between the original video $\mathbf{V}$ and the reconstructed video $\hat{\mathbf{V}}$ in the pixel space is calculated as the reconstruction loss:
\begin{equation}
\mathcal{L} = \frac{1}{|\mathcal{M}|}\sum_{m \in \mathcal{M}}|| \mathbf{V}(m) - \hat{\mathbf{V}}(m)||^2
\label{eq_loss_pretrain}
\end{equation}
where $\mathcal{M}$ is the set of masked positions.

\subsection{Downstream Fine-tuning}
\label{sec_finetuning}
After self-supervised pre-training, we then discard the lightweight decoder and only use the pre-trained high-capacity encoder for downstream fine-tuning (Fig. \ref{fig_general} (b)). The main difference in the encoder behavior between fine-tuning and pre-training is that we do not perform multi-scale fusion to aggregate the outputs of three stages (i.e., only use the high-level spatiotemporal representations in the last stage, as shown in Fig. \ref{fig_encoder}), as we observe worse results in our experiments. 

Based upon the pre-trained encoder, we apply global average pooling to the extracted spatiotemporal feature and append a fully connected network for final prediction.
We use different loss functions for different types of video-based facial affect analysis tasks. For the classification task, the cross-entropy loss is employed, i.e., 
\begin{equation}
\mathcal{L}_{\textrm{cls}} = -\sum_{k=1}^{K}{y_k \log{\hat{y}_k}}
\label{eq_loss_cls}
\end{equation}
where $\hat{\mathbf{y}} \in \mathbb{R}^{K}$ denotes the prediction, $\mathbf{y} \in \mathbb{R}^{K}$ is the target, and $K$ is the number of emotion classes.
For the regression task, we compute the mean square error between the prediction and the target: 
\begin{equation}
\mathcal{L}_{\textrm{reg}} = \frac{1}{K}|| \mathbf{y} - \hat{\mathbf{y}} ||^2
\label{eq_loss_reg}
\end{equation}
where $K$ is the number of emotion dimensions.

\section{Experiments}
\label{sec_exp}

\subsection{Implementation Details}
\label{sec_impl}

\textbf{TPSBT architecture.}
To meet different needs in real-world applications, we build two versions (i.e., base and small) of TPSBT as the encoder of SVFAP. For the small version TPSBT-S, $C=384$, $M_1=8$, $M_2=4$, $M_3=2$.
For the base version TPSBT-B, $C=512$, $M_1=12$, $M_2=6$, $M_3=3$.
The model size and computational costs of the small version are approximately half of the base version. Besides, for both versions, the spatial bottleneck token number $G$ is set to 8.

\begin{figure}[t]
	\centering
	\includegraphics[width=0.9\linewidth]{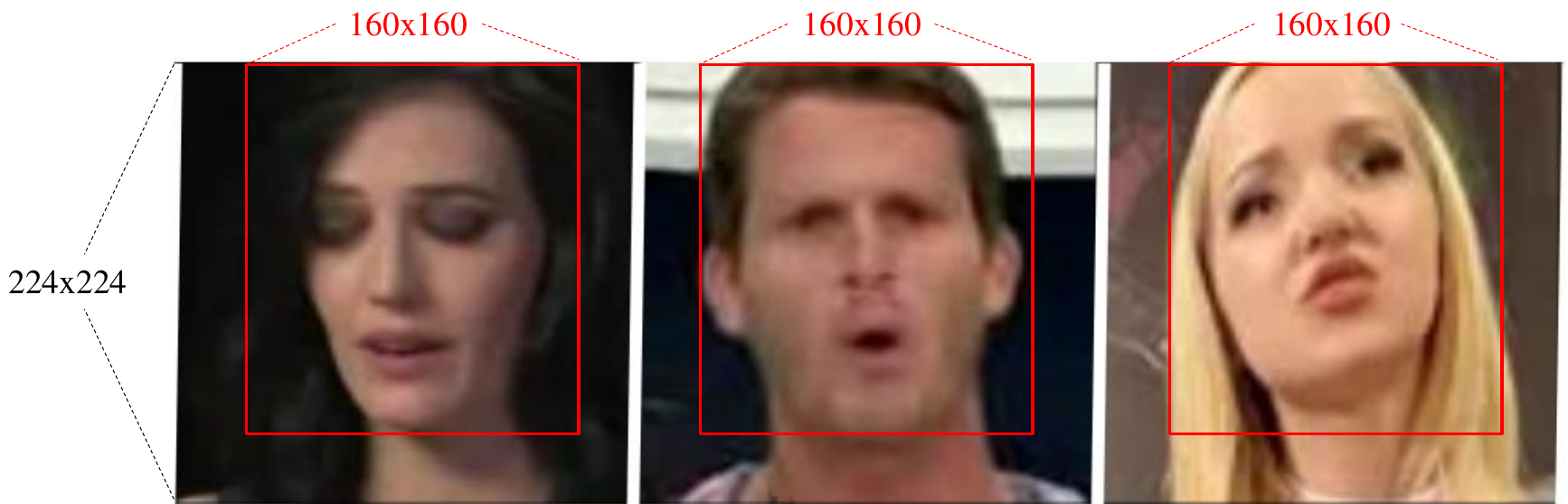}
	\caption{The illustration of the face patch location for VoxCeleb2 videos.}
	\label{fig_voxceleb2_face_patch}
\end{figure}

\textbf{Self-supervised pre-training.} 
% \subsubsection{Self-supervised Pre-training}
We pre-train TPSBT on a large-scale audio-visual dataset of human speech, VoxCeleb2 \cite{chung2018voxceleb2}. It includes more than 1 million video clips for over 6K celebrities extracted from about 150K videos uploaded to YouTube. We use its development set for self-supervised pre-training, which has 1,092,009 video clips from 145,569 videos. 
% The original video is with a resolution of $224\times224$. Since the speaker face typically does not cover the whole screen, we only use $160\times160$ patch located in the upper center of each video frame to filter irrelevant background information and reduce the computational costs.
As shown in Fig. \ref{fig_voxceleb2_face_patch}, the resolution of the original videos in VoxCeleb2 is $224\times224$, where faces typically appear in the upper-central part of the video. The remaining parts display the shoulders and neck, along with extraneous background information. Therefore, we only use a 160x160 patch from the upper-central location of the video, allowing us to remove irrelevant information while also reducing the model's input size, thus lowering computational costs.
We sample 16 frames from each video clip using a temporal stride of 4, resulting in $8\times10\times10$ input tokens after patch embedding when using the patch size of $2\times16\times16$.

We conduct experiments using the PyTorch framework with 4 Nvidia GeForce RTX 3090 GPUs. For the hyperparameter setting, we mainly follow VideoMAE \cite{tong2022videomae}. The main differences include the learning rate, batch size, and training epochs. Specifically, we adopt an AdamW optimizer with $\beta_1=0.9$ and $\beta_2=0.95$. The base learning rate is $3e-4$ and the weight decay is 0.05. The overall batch size is 256. We linearly scale the base learning rate with respect to the overall batch size, i.e., $\textrm{lr} = \textrm{base learning rate} \times \frac{\textrm{batch size}}{256}$. Besides, we employ a cosine decay learning rate scheduler. By default, we pre-train the model for 100 epochs with 5 warmup epochs and it takes about 5-6 days in our setting. 

\textbf{Downstream fine-tuning.}
% \subsubsection{Downstream Fine-tuning}
The input video clip size is also $16\times160\times160$ and the temporal stride is 4 in most cases. We adopt an AdamW optimizer with $\beta_1=0.9$ and $\beta_2=0.999$. The base learning rate is $1e-3$ and the overall batch size is 96. Other hyperparameters are the basically same as those in pre-training and can also refer to \cite{tong2022videomae} for more details. We fine-tune the pre-trained model for 100 epochs with 5 warmup epochs. For inference, we uniformly sample two clips along the temporal axis for each video sample and then compute the average score as the final prediction.

% \section{Results}

\subsection{Dynamic Facial Expression Recognition}

\subsubsection{Datasets}
We conduct experiments on 6 dynamic facial expression recognition datasets, including 3 large-scale in-the-wild datasets (i.e., DFEW \cite{jiang2020dfew}, MAFW \cite{liu2022mafw}, and FERV39k \cite{wang2022ferv39k}) and 3 small lab-controlled datasets (i.e., CREMA-D \cite{cao2014crema}, RAVDESS \cite{livingstone2018ryerson}, and eNTERFACE05 \cite{martin2006enterface}). We briefly introduce each of them below.

\textbf{DFEW} consists of 16,372 video clips which are extracted from more than 1,500 high-definition movies. 
Each video clip is annotated by 10 well-trained annotators with 7 basic emotions (i.e., anger, disgust, fear, happy, sad, surprise, and neutral). 
To align with previous studies \cite{jiang2020dfew, zhao2021former}, we evaluate the proposed method on 11,697 single-labeled clips using the default 5-fold cross-validation protocol.

\textbf{MAFW} is a multimodal compound affective dataset in the wild. It is composed of 10,045 video clips annotated with 11 compound emotions, including contempt, anxiety, helplessness, disappointment, and 7 basic emotions. 
In this paper, we only consider the video modality and conduct experiments on 9,172 single-labeled video clips. For evaluation, we follow the original paper \cite{liu2022mafw} to adopt the 5-fold cross-validation protocol.

\textbf{FERV39k} is currently the largest real-world dynamic facial expression recognition dataset. 
It includes 38,935 video clips which belong to 22 representative scenes in 4 different scenarios.
Each sample is annotated by 30 professional annotators with 7 basic emotions.
The whole dataset has been officially split into 80\% for training and the rest 20\% for test. 

\textbf{CREMA-D} is a high-quality audio-visual dataset for multimodal expression and perception of acted emotions. 
It consists of 7,442 video clips from 91 actors. 
Each video clip is labeled with 6 emotions, including happy, sad, anger, fear, disgust, and neutral. Since there is no official split for this dataset, we employ a 5-fold subject-independent cross-validation protocol.

\textbf{RAVDESS} is an audio-visual dataset of emotional speech and song. It consists of 2,880 video clips from 24 professional actors, each of which is labeled with 8 emotions (i.e., 7 basic emotions and calm). 
In this paper, we only use the speech part with 1,440 video clips. 
This dataset has no official split, we thus follow \cite{su2020msaf, fu2021cross} to adopt the same 6-fold subject-independent cross-validation protocol.

\textbf{eNTERFACE05} is also an audio-visual emotion dataset that contains about 1,200 video clips from more than 40 subjects. Each subject is asked to simulate six emotions, including anger, disgust, fear, happy, sad, and surprise. To make a fair comparison with previous work \cite{zhao2022spatial}, we employ a 5-fold subject-independent cross-validation protocol.

Following previous work \cite{jiang2020dfew, zhao2021former, liu2022mafw, wang2022ferv39k}, we use the weighted average recall (WAR, i.e., the  accuracy) and unweighted average recall (UAR, i.e., the mean class accuracy) as evaluation metrics for all datasets. 
Note that, for cross-validation, we aggregate the predictions and labels from all splits and then report the overall UAR and WAR.

\subsubsection{Ablation Studies}
\label{sec_ablation}
In this section, we conduct in-depth ablation experiments to investigate the impacts of several key factors in our proposed SVFAP. By default, we use TPSBT-B as the encoder. During downstream fine-tuning, all models share the same evaluation protocol. 
In addition to fine-tuning performance, we also show the number of model parameters and computational overhead measured in Floating Point Operations (FLOPs) \footnote{https://github.com/facebookresearch/fvcore} for efficiency comparison. 
Note that we denote pre-training FLOPs as FLOPs-P.
 Unless otherwise stated, all experiments are conducted on split 1 of DFEW.

\begin{figure}[t]
	\centering
	\includegraphics[width=0.9\linewidth]{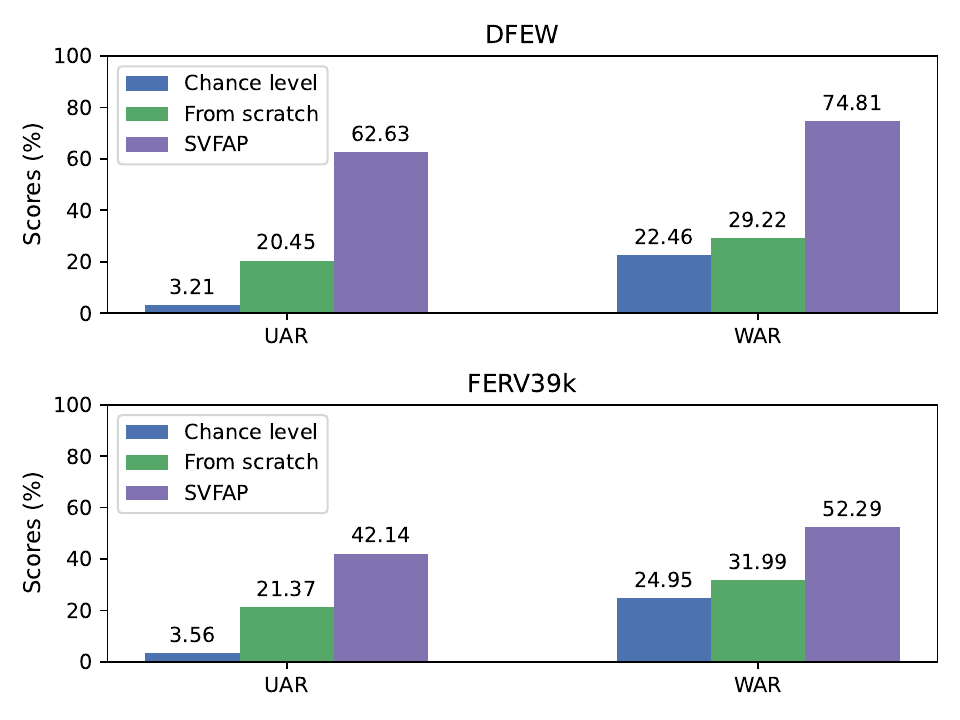}
	\caption{Ablation study of training from scratch.}
	\label{fig_ablation_from_scratch}
\end{figure}

\textbf{Training from scratch.} We first show the superiority of the proposed self-supervised pre-training method SVFAP by comparing it with training from scratch. 
Fig. \ref{fig_ablation_from_scratch} presents the comparison results on DFEW and FERV39k.
From the figure, we first observe that it is hard to achieve good results when training TPSBT-B from scratch. This observation is consistent with previous findings (i.e., vision Transformers are data-hungry) in computer vision \cite{dosovitskiy2020image, liu2021efficient}, and can be largely attributed to the lack of inductive bias and the small size of training datasets. Moreover, we find that our SVFAP significantly outperforms training from scratch, achieving about 42\% UAR and 45\% WAR improvements on DFEW, and 21\% UAR and 20\% WAR improvements on FERV39k. These encouraging results demonstrate that SVFAP provides an effective self-supervised pre-training mechanism for video-based facial affect analysis. We also notice that the performance gap between SVFAP and training from scratch on FERV39k is smaller than that on DFEW, probably due to the larger dataset size of the former (39K vs. 13K).

\begin{figure}[t]
	\centering
	\includegraphics[width=0.9\linewidth]{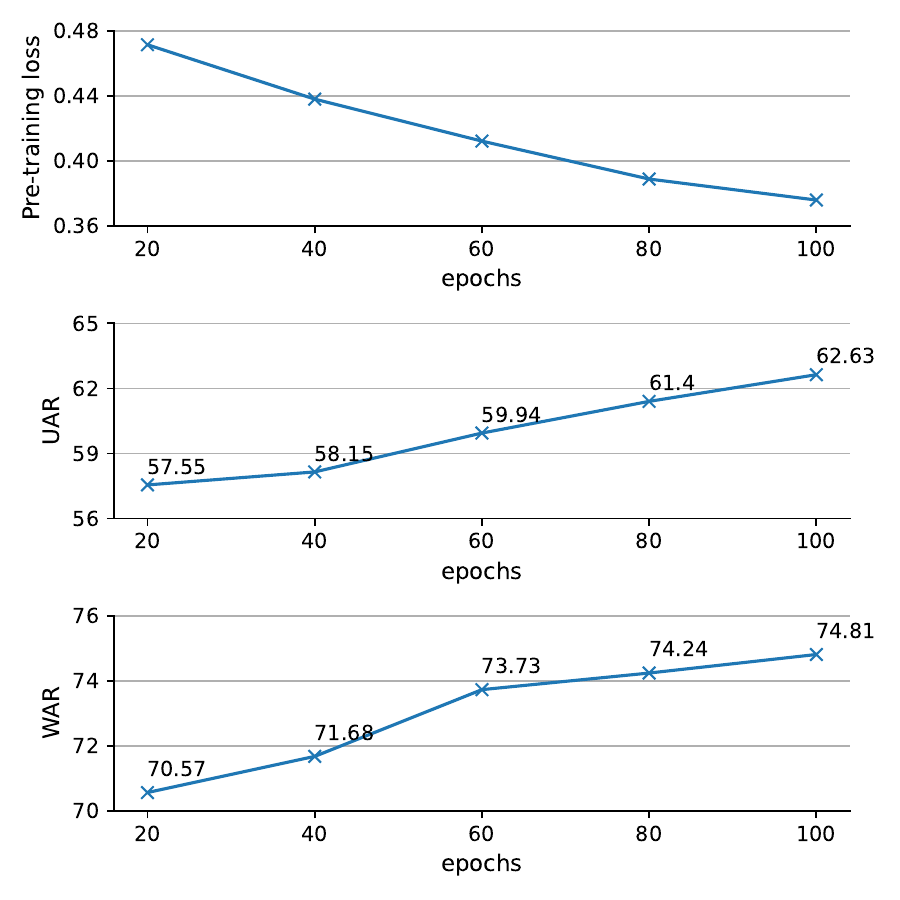}
	\caption{Ablation study of training schedule.}
	\label{fig_ablation_train_schedule}
\end{figure}

\begin{figure*}[t]
	\centering
	\includegraphics[width=1.0\linewidth]{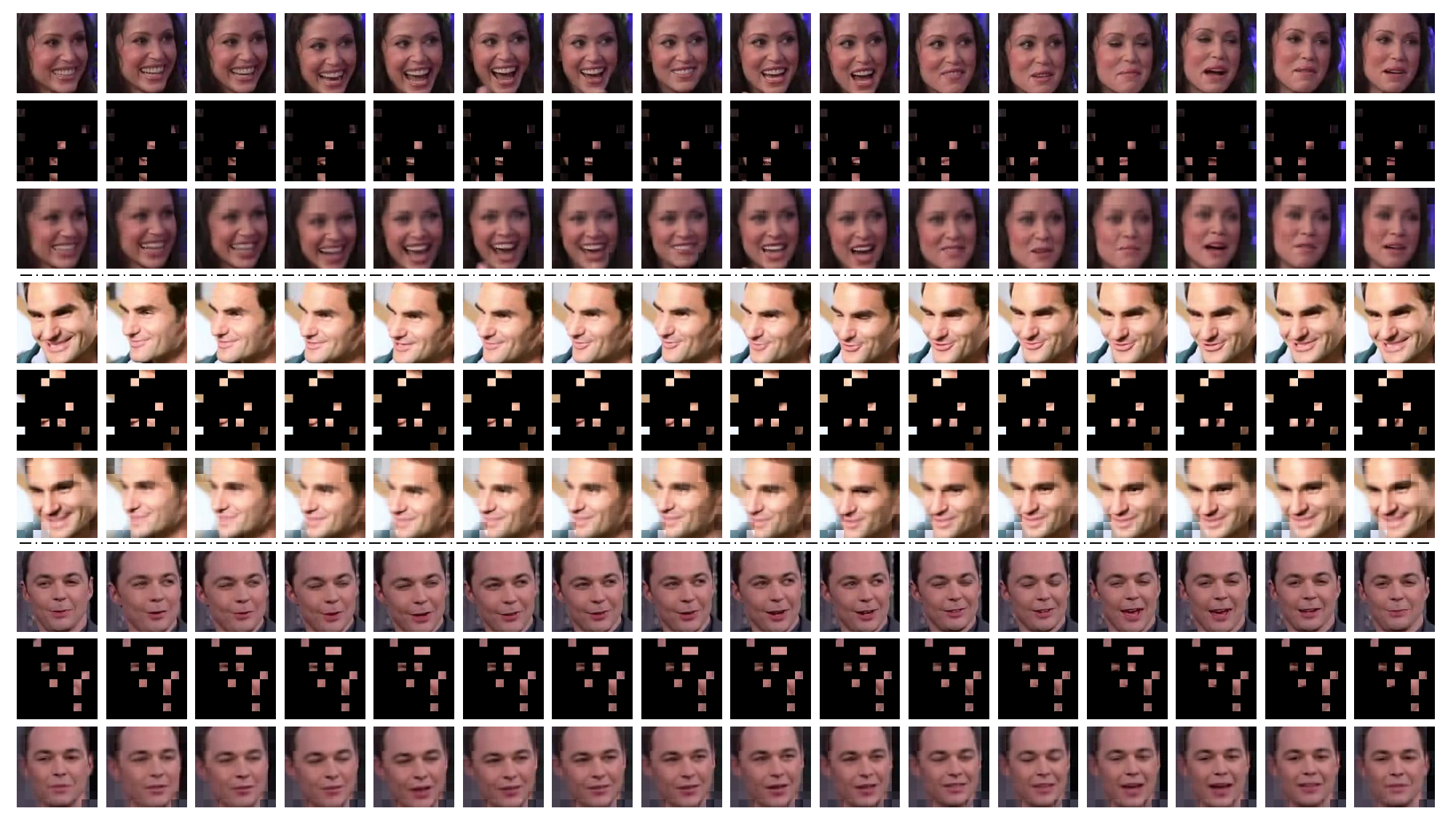}
	\caption{Reconstruction results of three randomly selected video samples from the test set of VoxCeleb2 with a masking ratio of 90\%. For each sample, we show the original video (top), masked input video (middle), and the reconstructed video (bottom).}
	\label{fig_recon}
\end{figure*}

\textbf{Training schedule.} We then explore the effect of training schedule length. As shown in Fig.  \ref{fig_ablation_train_schedule}, we find that as the pre-training process goes on, the pre-training loss decreases steadily and the downstream fine-tuning performance improves consistently. This finding is in line with previous works on self-supervised learning \cite{he2022masked, feichtenhofer2022masked, tong2022videomae}. It also should be noted that we do not observe clear performance saturation when reaching default maximum epochs (i.e., 100), which indicates that longer pre-training could further improve model performance in downstream fine-tuning. However, due to limited computational resources, we leave it for future work and hope the community can conduct follow-up studies.

% (add reconstruction figure)
% Moreover, we visualize several reconstructed video samples using our best pre-trained model in Fig. \ref{fig_recon}. Note that these samples are randomly selected from the test set of VoxCeleb2 thus can not be seen by our model during pre-training. We find that even under such a high masking ratio (i.e., 90\%), SVFAP still can generate satisfying reconstruction results, especially for dynamic facial expressions. This indicates that, benefited from the challenging masked facial video autoencoding task, our model can reason over high-level and meaningful spatiotemporal semantics from limited visible input to recover masked information.
Moreover, we visualize several reconstructed video samples using our best pre-trained model in Fig. \ref{fig_recon}. Note that these samples are randomly selected from the test set of VoxCeleb2, whose speakers are disjoint with those from the development set used for pre-training. Thus, they are not seen by our model during pre-training. We find that even under such a high masking ratio (i.e., 90\%), SVFAP still can generate satisfying reconstruction results, especially for dynamic facial expressions. This indicates that, benefiting from the challenging masked facial video autoencoding task, our model can reason over high-level and meaningful spatiotemporal semantics from limited visible input to recover masked information.

\textbf{Dataset scale.} We investigate how the model behaves when pre-trained with different dataset scales. For this purpose, we randomly generate a range of subsets with increasing sizes from the whole dataset, i.e., 1\%, 5\%, 10\%, and 20\%. Note that we proportionally increase the pre-training epochs for these subsets to ensure the same pre-training cost. The results are shown in Table \ref{tab_ablation_data_scale}. We can observe that larger pre-training datasets generally lead to better fine-tuning results. This is expected as more diversified training samples typically result in better generalization. It is also worth noting that, even with 11K unlabeled data, our method still achieves promising performance, which demonstrates that our SVFAP is a data-efficient self-supervised video facial affect learner.

\begin{table}[]
\caption{Ablation study of pre-training dataset scale.}
\label{tab_ablation_data_scale}
\centering
\begin{tabular}{lcccc}
\toprule
Percentage & Size    & Epochs   & UAR    & WAR  \\
\midrule
1\%       & 11K      & 10000     & 57.65  & 70.61     \\
5\%       & 55K      &  2000     & 60.34  & 73.00     \\
10\%      & 110K     &  1000     & 61.09  & 73.56     \\
20\%      & 220K     &   500     & 61.54  & 73.76      \\
100\%     & 1.1M     &   100     & \textbf{62.63}  & \textbf{74.81}     \\
\bottomrule
\end{tabular}
\end{table}

\begin{table}[]
\caption{Ablation study of the masking ratio. FLOPs-P: Pre-training FLOPs.}
\label{tab_ablation_masking_ratio}
\centering
\begin{tabular}{ccccccc}
\toprule
Percentage & \tabincell{c}{\#Params\\(M)} & \tabincell{c}{FLOPs\\(G)} & \tabincell{c}{FLOPs-P\\(G)}    & UAR    & WAR  \\
\midrule
75\%       &  77.6    & 43.6  & 18.4     & 60.89  & 73.91     \\
85\%       &  77.6    & 43.6  & 14.7     & 61.87  & 74.77     \\
90\%       &  77.6    & 43.6  & 12.9     & \textbf{62.63}  & \textbf{74.81}      \\
95\%       &  77.6    & 43.6  & 11.2     & 59.93  & 73.17      \\
\bottomrule
\end{tabular}
\end{table}

\textbf{Masking ratio.} The influence of the masking ratio is presented in Table \ref{tab_ablation_masking_ratio}. We can find that the masking ratio of 90\% has the best performance. The lower masking ratios of 75\% and 85\% achieve worse performance, although the encoder accepts more video tokens as input and has higher computational costs during pre-training. A higher masking ratio of 95\% results in lower pre-training cost, however, the performance degrades significantly. These results are consistent with previous findings \cite{tong2022videomae, feichtenhofer2022masked}. Therefore, we set 90\% as the default masking ratio.

% with pre-training FLOPs
\begin{table}[]
\caption{Ablation study of model variants. TP: Temporal Pyramid. SBT: Spatial Bottleneck Transformer. FLOPs-P: Pre-training FLOPs.}
\label{tab_ablation_model_variant}
\centering
% \resizebox{\linewidth}{!}{
\begin{tabular}{ccccccc}
\toprule
TP &  SBT &  \tabincell{c}{\#Params\\(M)} & \tabincell{c}{FLOPs\\(G)} & \tabincell{c}{FLOPs-P\\(G)}   & UAR    & WAR  \\
\midrule
$\times$         & $\times$            &  76.4    & 76.9  & 14.9  & 61.71  & 74.41     \\
\checkmark       & $\times$            &  77.5    & 53.1  & 13.1  & 60.84  & 74.28     \\
$\times$         & \checkmark          &  76.6    & 49.9  & 13.5  & \textbf{62.66}  & \textbf{75.02}     \\
\checkmark       & \checkmark          &  77.6    & 43.6  & 12.9  & 62.63  & 74.81     \\
\bottomrule
\end{tabular}
% }
\end{table}

\textbf{Model variants.} We ablate different model variants of TPSBT, including 1) no TP and SBT, i.e., the vanilla ViT baseline, by removing temporal pyramid learning modules and replacing SBT with a standard Transformer of similar size in the last two stages. 2) only TP, by replacing SBT with a standard Transformer of similar size. 3) only SBT, by removing temporal pyramid learning modules. As presented in Table \ref{tab_ablation_model_variant}, we have the following observations: 1) When compared with the vanilla ViT baseline, TP largely reduces computational overhead (about 30\% and 14\% FLOPs reduction for fine-tuning and pre-training respectively) and achieves comparable performance, with only 1.1M additional parameters. 2) SBT not only reduces large computations (about 35\% FLOPs and 11\% FLOPs-P reduction) but also achieves the best performance among all variants. 3) When combining TP and SBT together, our default full model achieves the lowest computational costs while almost maintaining the best performance of SBT. Specifically, TPSBT significantly reduces about 43\% and 15\% FLOPs during fine-tuning and pre-training and outperforms the vanilla ViT baseline by 0.92\% UAR and 0.40\% WAR, with the sacrifice of a slight increase of model parameters (1.2M). These results indicate that temporal pyramid learning and spatial bottleneck mechanism can greatly remove redundant spatiotemporal information in 3D facial videos (from temporal and spatial perspectives respectively) and help the model to concentrate on the informative one, thus contributing to lower computational costs and better model performance.

\textbf{Multi-scale fusion.} We explore the effect of multi-scale spatiotemporal feature fusion during both self-supervised pre-training and downstream fine-tuning. As shown in Table \ref{tab_ablation_multiscale_fusion}, we find that employing multi-scale fusion during pre-training achieves better results than those that do not use it, which shows that integrating spatiotemporal features in different levels for decoder reconstruction can help the encoder to learn more useful representations. However, we observe slight performance degradation when using it during fine-tuning. This result could be partly ascribed to the simple and parameter-free temporal upsampling method (i.e., nearest-neighbor interpolation) used for multi-scale fusion. 
Another reason is that, compared to low-level details, high-level emotional semantics are more crucial for downstream affect analysis tasks.
Therefore, we only use the high-level feature from the last stage of the encoder during fine-tuning.

\begin{table}[]
\caption{Ablation study of multi-scale fusion.}
\label{tab_ablation_multiscale_fusion}
\centering
\begin{tabular}{cccc}
\toprule
Pre-training &  Fine-tuning   & UAR    & WAR  \\
\midrule
$\times$     & $\times$       & 61.78  & 74.43     \\
$\times$     & \checkmark     & 61.24  & 74.17     \\
\checkmark   & $\times$       & \textbf{62.63}  & \textbf{74.81}    \\
\checkmark   & \checkmark     & 62.39  & 74.74    \\
\bottomrule
\end{tabular}
\end{table}

\begin{table}[]
\caption{Ablation study of spatial bottleneck tokens. FLOPs-P: Pre-training FLOPs.}
\label{tab_ablation_sbt_tokens}
\centering
\begin{tabular}{cccccc}
\toprule
Number          &  \tabincell{c}{\#Params\\(M)} & \tabincell{c}{FLOPs\\(G)} & \tabincell{c}{FLOPs-P\\(G)}   & UAR    & WAR  \\
\midrule
4                &  77.6     & 43.2  & 12.6  & 61.76  & 74.08    \\
8                &  77.6     & 43.6  & 12.9  & 62.63  & \textbf{74.81}    \\
16               &  77.6     & 44.4  & 13.7  & \textbf{62.86}  & 74.69    \\
\bottomrule
\end{tabular}
\end{table}

\textbf{Spatial bottleneck tokens.} SBT utilizes spatial attention to generate several global semantic bottleneck tokens for spatial redundancy elimination and computational cost reduction, thus it is necessary to investigate how the number of spatial bottleneck tokens influence model performance. Table \ref{tab_ablation_sbt_tokens} presents the ablation results. We see that too few tokens (i.e., 4) hurt the model performance as it might be too aggressive to perform spatial compression and incur critical information loss. Besides, too many tokens (i.e., 16) increase model computational costs but do not contribute to significantly better results. Therefore, we set the spatial bottleneck token number to 8 by default. 

% with pre-training FLOPs
\begin{table}[]
\caption{Ablation study of temporal downsampling. FLOPs-P: Pre-training FLOPs.}
\label{tab_ablation_ttemporal_downsampling}
\centering
\begin{tabular}{cccccc}
\toprule
Type           &  \tabincell{c}{\#Params\\(M)} & \tabincell{c}{FLOPs\\(G)} & \tabincell{c}{FLOPs-P\\(G)}   & UAR    & WAR  \\
\midrule
Avg     &  76.6      & 43.3  & 12.9  & 62.04  & 74.42    \\
Max     &  76.6      & 43.3  & 12.9  & 62.25  & 74.16    \\
Conv    &  77.6      & 43.6  & 12.9  & \textbf{62.63}  & \textbf{74.81}    \\
\bottomrule
\end{tabular}
\end{table}

\textbf{Temporal downsampling.} We explore the effect of three different methods for temporal downsampling in temporal pyramid learning, including simple average pooling, max pooling, and the default strided convolution. As presented in Tab. \ref{tab_ablation_ttemporal_downsampling}, we find that the convolution-based method outperforms the other two pooling methods with the sacrifice of slightly more model parameters and negligible FLOPs increase, which justifies our default design choice.

\begin{table*}[]
\caption{Comparison with state-of-the-art supervised and self-supervised pre-trained models on split 1 of DFEW. SSL: Self-Supervised Learning or not.}
\label{tab_pretrained}
\centering
% \resizebox{\linewidth}{!}{
\begin{tabular}{lccccccccc}
% \begin{tabular}{l|l|l|l|l|l|l|l|l|l}
\toprule
Method      & SSL & Pre-training Dataset                & Dataset Type        & Architecture & \tabincell{c}{\#Params\\(M)} & \tabincell{c}{FLOPs\\(G)} & UAR   & WAR   \\
\midrule
TimeSformer \cite{bertasius2021space} & $\times$        & ImageNet-21K+Kinetics-400        & Image+Video & TimeSformer  & 121 &  198  & 58.90 & 71.48 \\
TimeSformer \cite{bertasius2021space} & $\times$        & \tabincell{c}{ImageNet-21K+Kinetics-400\\+HowTo100M} & Image+Video & TimeSformer  & 121 &  198  & 59.13 & 71.62 \\ 
MViT \cite{fan2021multiscale}       & $\times$       & Kinetics-400                        & Video       & MViT-B       & 53  &  45  & 54.19 & 65.27 \\
MViT \cite{fan2021multiscale}       & $\times$       & Kinetics-600                        & Video       & MViT-B       & 53  &  45 & 55.80 & 66.85 \\
MViTv2 \cite{li2022mvitv2}      & $\times$       & Kinetics-400                            & Video       & MViTv2-B     & 51  &  42 & 55.12 & 67.88 \\
Video Swin \cite{liu2022video}  & $\times$       & ImageNet-1K+Kinetics-400                        & Image+Video & Swin-S       & 50  &  55 & 57.13 & 69.96 \\
Video Swin \cite{liu2022video}  & $\times$       & ImageNet-21K+Kinetics-400                       & Image+Video & Swin-B       & 88     &  93   & 59.38 & 71.90 \\
\midrule
FAb-Net \cite{wiles2018self}    & \checkmark     & VoxCeleb1\&2                            & Video       & ConvNet   & 6   &  33 & 45.52 & 56.01 \\
TCAE \cite{li2019self}          & \checkmark     & VoxCeleb1\&2                            & Video       & ConvNet   & 6   &  33 & 45.29 & 56.35 \\
FaceCycle \cite{chang2021learning}   & \checkmark     & VoxCeleb1\&2                       & Video       & ConvNet   & 4   &  42 & 42.27 & 54.05 \\
BMVC’20 \cite{lu2020self}      & \checkmark     & VoxCeleb2                                & Video       & ResNet-18     & 12  &  15 & 56.55 & 67.12 \\
MoCo \cite{zhao2020makes}      & \checkmark     & CelebA                                   & Image       & ResNet-50     & 32  &  34 & 53.47 & 67.45 \\
SwAV \cite{bulat2022pre}      & \checkmark     & VGGFace2                                  & Image       & ResNet-50     & 32    &  34 & 58.18 & 68.99 \\
$\rho$BYOL \cite{feichtenhofer2021large}  & \checkmark     & Kinetics-400                          & Video       & SlowOnly-R50        & 32  &  43 & 58.60 & 69.81 \\
SVT \cite{ranasinghe2022self}  & \checkmark     & Kinetics-400                          & Video       & TimeSformer        & 121  &  198 & 57.07 & 70.01 \\
VideoMAE \cite{tong2022videomae}  & \checkmark     & Kinetics-400                          & Video       & ViT-B        & 86  &  81 & 58.32 & 70.94 \\
FaRL \cite{zheng2022general}  & \checkmark     & LAION-FACE                                & Image+Text  & ViT-B        & 93  & 141 & 58.91 & 72.15 \\
\midrule
SVFAP-S (ours)  & \checkmark     & VoxCeleb2                   & Video       & TPSBT-S & 30  & 18  & 59.70 & 72.70 \\ 
SVFAP-B (ours)  & \checkmark     & VoxCeleb2                   & Video       & TPSBT-B & 78  & 44  & \textbf{62.63} & \textbf{74.81} \\ 
\bottomrule
\end{tabular}
% }
\end{table*}

\begin{figure*}[t]
	\centering
	\includegraphics[width=1.0\linewidth]{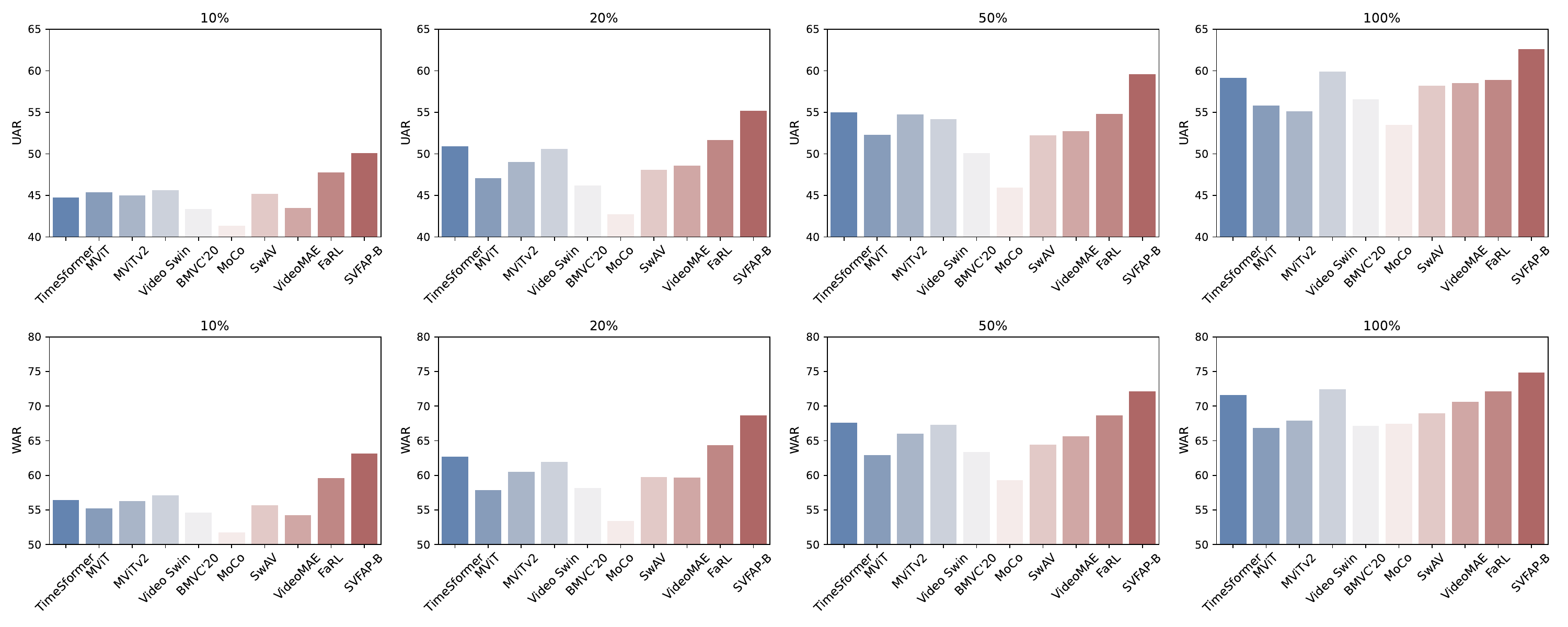}
	\caption{Comparisons with state-of-the-art supervised and self-supervised pre-trained models on the split 1 of DFEW in the few-shot setting.}
	\label{fig_dfew_few_shot}
\end{figure*}

\begin{table*}[]
\caption{Comparison with state-of-the-art methods on DFEW.}
\label{tab_dfew_sota}
\centering
% \resizebox{\linewidth}{!}{
\begin{tabular}{lccccccccccc}
\toprule
\multirow{2}{*}{Method} & \multirow{2}{*}{\tabincell{c}{\#Params\\(M)}}  & \multirow{2}{*}{\tabincell{c}{FLOPs\\(G)}}  
& \multicolumn{7}{c}{Accuracy of Each Emotion (\%)}  & \multicolumn{2}{c}{Metric (\%)} \\ \cmidrule(lr){4-10} \cmidrule(lr){11-12} 
&                            &                       & Happy          & Sad            & Neutral        & Anger          & Surprise       & Disgust       & Fear           & UAR             & WAR            \\ \midrule

3D ResNet-18 \cite{hara2018can}                     &  -  &  8  & 76.32 & 50.21 & 64.18 & 62.85 & 47.52 & 0.00 & 24.56 & 46.52 & 58.27    \\ 
EC-STFL \cite{jiang2020dfew}                        &  -  &  8  & 79.18 & 49.05 & 57.85 & 60.98 & 46.15 & 2.76 & 21.51 & 45.35 & 56.51    \\ 
ResNet-18+LSTM \cite{zhao2021former}                &  -  &  8  & 83.56 & 61.56 & 68.27 & 65.29 & 51.26 & 0.00 & 29.34 & 51.32 & 63.85    \\ 
ResNet-18+GRU  \cite{zhao2021former}                &  -  &  8  & 82.87 & 63.83 & 65.06 & 68.51 & 52.00 & 0.86 & 30.14 & 51.68 & 64.02    \\ 
Former-DFER \cite{zhao2021former}                   &  18 &  9  & 84.05 & 62.57 & 67.52 & 70.03 & 56.43 & 3.45 & 31.78 & 53.69 & 65.70   \\ 
CEFLNet \cite{liu2022clip}                          &  13 & -   & 84.00 & 68.00 & 67.00 & 70.00 & 52.00 & 0.00 & 17.00 & 51.14 & 65.35   \\ 
EST \cite{liu2021expression}                        &  43 &  -  & 86.87 & 66.58 & 67.18 & 71.84 & 47.53 & 5.52 & 28.49 & 53.43 & 65.85   \\
STT \cite{ma2022spatio}                             &  -  &  -  & 87.36 & 67.90 & 64.97 & 71.24 & 53.10 & 3.49 & 34.04 & 54.58 & 66.65   \\ 
DPCNet \cite{wang2022dpcnet}                        &  51 & 10  & -     & -     & -     & -     & -     & -    & -     & 57.11 & 66.32   \\ 
NR-DFERNet \cite{li2022nr}                          &  -  &  6  & 88.47 & 64.84 & 70.03 & 75.09 & 61.60 & 0.00 & 19.43 & 54.21 & 68.19   \\ 
IAL \cite{li2023intensity}                      &  -  &  -  & 87.95 & 67.21 & 70.10 & 76.06 & 62.22 & 0.00 & 26.44 & 55.71 & 69.24   \\
M3DFEL \cite{wang2023rethinking}         &  -   &  2  & 89.59 & 68.38 & 67.88 & 74.24 & 59.69 & 0.00 & 31.63 & 56.10 & 69.25       \\
\midrule
SVFAP-S (ours)     &  30   & 18  & 92.39 & 74.92 & 70.40 & 76.90 & 62.70 & 8.28 & 37.58 & 60.45  & 72.67 \\ 
SVFAP-B (ours)     &  78   &  44  & \textbf{93.13} & \textbf{76.98} & \textbf{72.31} & \textbf{77.54} & \textbf{65.42} & \textbf{15.17} & \textbf{39.25} & \textbf{62.83}  & \textbf{74.27} \\ 
\bottomrule
\end{tabular}
% }
\end{table*}

\begin{table*}[]
\caption{Comparison with state-of-the-art methods on FERV39k.}
\label{tab_ferv39k_sota}
\centering
% \resizebox{\linewidth}{!}{
\begin{tabular}{lccccccccccc}
\toprule
\multirow{2}{*}{Method} & \multirow{2}{*}{\tabincell{c}{\#Params\\(M)}}  & \multirow{2}{*}{\tabincell{c}{FLOPs\\(G)}}  
& \multicolumn{7}{c}{Accuracy of Each Emotion (\%)}  & \multicolumn{2}{c}{Metric (\%)} \\ \cmidrule(lr){4-10} \cmidrule(lr){11-12} 
                        &  &  & Happy          & Sad            & Neutral        & Anger          & Surprise       & Disgust       & Fear           & UAR             & WAR            \\ \midrule
C3D \cite{tran2015learning}          &  78  &  -   & 48.20   & 35.53   & 52.71     & 13.72   & 3.45       & 4.93     & 0.23    & 22.68                       & 31.69                      \\
P3D \cite{qiu2017learning}           &  -   &  -   & 61.85   & 42.21   & 49.80     & 42.57   & 10.50      & 0.86     & 5.57    & 30.48                       & 40.81                      \\
R(2+1)D \cite{tran2018closer}        &  -   &  -   & 59.33   & 42.43   & 50.82     & 42.57   & 16.30      & 4.50     & 4.87    & 31.55                       & 41.28                      \\
3D ResNet-18 \cite{hara2018can}       &  33  &  -   & 57.64   & 28.21   & 59.60     & 33.29   & 4.70       & 0.21     & 3.02    & 26.67                       & 37.57                      \\
ResNet-18+LSTM \cite{wang2022ferv39k} &  -   &  -   & 61.91   & 31.95   & 61.70     & 45.93   & 14.26      & 0.00     & 0.70    & 30.92                       & 42.59                      \\
VGG-13+LSTM \cite{wang2022ferv39k}    &  -   &  -   & 66.26   & 51.26   & 53.22     & 37.93   & 13.64      & 0.43     & 4.18    & 32.42                       & 43.37                      \\
Two C3D \cite{wang2022ferv39k}       &  -   &  -   & 54.85   & 52.91   & 60.67     & 31.34   & 5.96       & 2.36     & 6.96    & 30.72                       & 41.77                      \\
Two ResNet-18+LSTM \cite{wang2022ferv39k}     &  -   &  -   & 59.00   & 45.87   & 61.90     & 40.15   & 9.87       & 1.71     & 0.46    & 31.28                       & 43.20                      \\
Two VGG-13+LSTM \cite{wang2022ferv39k}  &  -   &  -   & 69.65   & 47.31   & 52.55     & 47.88   & 7.68       & 1.93     & 2.55    & 32.79                       & 44.54                      \\
Former-DFER \cite{zhao2021former}    &  18  &  9   & 65.65   & 51.33   & 56.74     & 43.64   & 21.94      & 8.57     & 12.53   & 37.20                       & 46.85                 \\
STT \cite{ma2022spatio}              &  -   &  -   & 69.77   & 47.81   & 59.14     & 47.41   & 20.22      & 10.49    & 9.51    & 37.76           & 48.11                      \\
NR-DFERNet \cite{li2022nr}           &  -   &  6   & 69.18   & \textbf{54.77}   & 51.12     & 49.70   & 13.17      & 0.00     & 0.23    & 33.99                       & 45.97                      \\
IAL \cite{li2023intensity}       &  -   &  -   &  -      &  -      &  -        &  -      &  -         &  -       &  -      & 35.82                       & 48.54          \\ 
M3DFEL \cite{wang2023rethinking}         &  -   &  2  &  -      &  -      &  -        &  -      &  -         &  -       &  -        & 35.94 & 47.67       \\
\midrule
SVFAP-S (ours)                       &  30  &  18  & \textbf{75.02}   & 52.12   & 61.34     & 48.69   & 23.04      & 12.85    &  15.31  & 41.19                       & 51.34                      \\ 
SVFAP-B (ours)                       &  78  &  44  & 74.00   & 53.34   & \textbf{62.26}     & \textbf{51.11}   & \textbf{25.24}      & \textbf{13.28}    &  \textbf{15.78}  & \textbf{42.14}                       & \textbf{52.29}             \\ 
\bottomrule
\end{tabular}
%}
\end{table*}

\begin{table*}[]
\caption{Comparison with state-of-the-art methods on MAFW. AN: Anger. DI: Disgust. FE: Fear. HA: Happiness. NE: Neutral. SA: Sadness. SU: Surprise. CO: Contempt. AX: Anxiety. HL: Helplessness. DS: Disappointment.}
\label{tab_mafw_sota}
\centering
\resizebox{\linewidth}{!}{
\begin{tabular}{lccccccccccccccc}
\toprule
\multirow{2}{*}{Method} & \multirow{2}{*}{\tabincell{c}{\#Params\\(M)}}  & \multirow{2}{*}{\tabincell{c}{FLOPs\\(G)}}  
& \multicolumn{11}{c}{Accuracy of Each Emotion (\%)}  & \multicolumn{2}{c}{Metric (\%)} \\ 
\cmidrule(lr){4-14} \cmidrule(lr){15-16} 
&  &  & AN    & DI    & FE    & HA    & NE    & SA    & SU    & CO   & AX    & HL   & DS   & UAR  & WAR  \\ 
\midrule
ResNet-18 \cite{he2016deep}        &  11  &  -  & 45.02 & 9.25  & 22.51 & 70.69 & 35.94 & 52.25 & 39.04 & 0.00 & 6.67  & 0.00 & 0.00 & 25.58 & 36.65          \\
ViT \cite{dosovitskiy2020image}   &  86  &  -  & 46.03 & 18.18 & 27.49 & 76.89 & 50.70 & 68.19 & 45.13 & 1.27 & 18.93 & 1.53 & 1.65 & 32.36 & 45.04          \\
C3D \cite{tran2015learning}       &  78 &  -  & 51.47 & 10.66 & 24.66 & 70.64 & 43.81 & 55.04 & 46.61 & 1.68 & 24.34 & 5.73 & 4.93 & 31.17 & 42.25          \\
ResNet-18+LSTM \cite{liu2022mafw} &  -  &  -  & 46.25 & 4.70  & 25.56 & 68.92 & 44.99 & 51.91 & 45.88 & 1.69 & 15.75 & 1.53 & 1.65 & 28.08           & 39.38          \\
ViT+LSTM \cite{liu2022mafw}       &  -  &  -  & 42.42 & 14.58 & \textbf{35.69} & 76.25 & 54.48 & 68.87 & 41.01 & 0.00 & 24.40 & 0.00 & 1.65 & 32.67           & 45.56          \\
C3D+LSTM \cite{liu2022mafw}       &  -  &  -  & 54.91 & 0.47  & 9.00  & 73.43 & 41.39 & 64.92 & 58.43 & 0.00 & 24.62 & 0.00 & 0.00 & 29.75           & 43.76          \\
Former-DFER \cite{zhao2021former}    &  18  &  9  &  -      &  -      &  -        &  -      &  -         &  -       &  -   &  -      &  -      &  -        &  -       & 31.16	     & 43.27     \\
T-ESFL \cite{liu2022mafw}        &  -  &  -  & 62.70 & 2.51  & 29.90 & 83.82 & 61.16 & 67.98 & 48.50 & 0.00 & 9.52  & 0.00 & 0.00 & 33.28 & 48.18     \\
\midrule
SVFAP-S (ours)                   &  30 &  18 & 63.88 & 19.56 & 30.88 & \textbf{84.46} & \textbf{62.83} & 68.37 & \textbf{59.61} & 1.27 & 31.88 & 7.63 &  7.69 & 39.82 & 53.89        \\
SVFAP-B (ours)                   &  78 &  44 & \textbf{64.60} & \textbf{25.20} & 35.68 & 82.77 & 57.12 & \textbf{70.41} & 58.58 & \textbf{8.05} & \textbf{32.42} & \textbf{8.40} & \textbf{9.89} & \textbf{41.19} & \textbf{54.28}        \\
\bottomrule
\end{tabular}
}
\end{table*}

\subsubsection{Comparison with Previous Pre-trained Models}
In this section, we show the effectiveness of the proposed self-supervised learning method by comparing it with previous state-of-the-art supervised and self-supervised pre-trained models. 
The supervised models contain four advanced video Transformers pre-trained on large-scale labeled video or image datasets, including TimeSformer \cite{bertasius2021space}, MViT \cite{fan2021multiscale}, MViTv2 \cite{li2022mvitv2}, and Video Swin Transformer \cite{liu2022video}.
The self-supervised part involves eight cutting-edge models pre-trained on massive unlabeled data (most are facial images or videos), including FAb-Net \cite{wiles2018self}, TCAE \cite{li2019self}, FaceCycle \cite{chang2021learning}, BMVC’20 \cite{lu2020self}, MoCo \cite{zhao2020makes}, SwAV \cite{bulat2022pre}, $\rho$BYOL \cite{feichtenhofer2021large}, SVT \cite{ranasinghe2022self}, VideoMAE \cite{tong2022videomae}, and FaRL \cite{zheng2022general}.
It should be noted, all self-supervised models except $\rho$BYOL, SVT, and VideoMAE are 2D models, i.e., they can not process video inputs directly. Therefore, we add a standard Transformer block on top of them to enable temporal sequential modeling. Besides, all models share the same evaluation protocol to ensure a fair comparison.

The comprehensive comparison results on the split 1 of DFEW are presented in Table \ref{tab_pretrained}. 
From the table, we have the following key observations:
\begin{itemize}
\item Our SVFAP outperforms all supervised pre-trained video Transformers. Specifically, SVFAP-B surpasses the best-performing Video Swin Base (Swin-B) model by 3.25\% UAR and 2.91\% WAR, although Swin-B has a larger model size (87.6M vs. 77.6M) and more than double FLOPs (92.5G vs. 43.6G), and is pre-trained on the combination of large-scale labeled images and videos. The more encouraging thing is that even our small model SVFAP-S still outperforms Swin-B slightly, thus amply demonstrating the remarkable superiority of our proposed method. We also notice that TimeSformer also achieves satisfying results but it suffers from huge computational costs and has much more parameters. 
To sum up, the comparison results with supervised models show that our method can learn strong and transferable affect-related facial representations from large-scale video data without using any human-annotated labels.
\item Compared with self-supervised pre-trained models, SVFAP also achieves the best performance. Notably, when using the same Voxceleb2 dataset for pre-training, both our base and small model improve FAb-Net, TCAE, FaceCycle, and BMVC'20, by a significant margin, which indicates that masked facial video autoencoding is an effective task for self-supervised pre-training. Our method also beats contrastive learning-based methods (i.e., MoCo, SwAV, and SVT), verifying the advantage of the generative paradigm in self-supervised learning. 
When compared to VideoMAE, SVFAP-S still shows better performance while having 2.9$\times$ fewer parameters and 4.5$\times$ fewer FLOPs. 
Finally, our method also outperforms the best-performing FaRL (i.e., 3.72\% UAR and 2.66\% WAR improvements for SVFAP-B, 0.32\% UAR and 0.80\% WAR improvements for SVFAP-S), although FaRL has much more FLOPs and parameters and is pre-trained on a huge multimodal dataset. 
To summarize, the above comparison results with self-supervised models show that our SVFAP is an effective and efficient self-supervised video facial affect perceiver. 
% \end{enumerate}
\end{itemize}

In addition to the evaluation in full data regime, we also conduct experiments to investigate the generalization ability of SVFAP under few-shot settings. To this end, we randomly select 50\%, 20\%, and 10\% samples from the training set of DFEW split 1 to obtain a series of new training sets while keeping the test set unchanged. For simplicity, we only choose several representative models in Table \ref{tab_pretrained} for evaluation. Note that, for the method with more than one model, we only use the best one. The results are reported in Fig. \ref{fig_dfew_few_shot}. We can observe that: 1) As expected, the performance of all methods degrades accordingly when fewer training samples are used. 2) SVFAP consistently outperforms all compared methods under each few-shot setting, which verifies the strong adaptation ability of the proposed methods in the low data regime. This ability is particularly important considering the longstanding data scarcity issue in affective computing. Notably, even with 10\% (about 935 samples) training data, our method still achieves promising results (more than 50\% UAR and 63\% WAR). 

\subsubsection{Comparisons with State-of-the-art Methods}

In this section, we compare the proposed method with state-of-the-art methods on both in-the-wild and lab-controlled dynamic facial expression recognition datasets. 

We first show the comparison results on three large in-the-wild datasets, including DFEW, FERV39k, and MAFW. 
The results on three datasets are reported in Table \ref{tab_dfew_sota}, \ref{tab_ferv39k_sota}, and \ref{tab_mafw_sota}, respectively.
The comparison baselines can be roughly divided into three categories: 
1) the combination of convolution neural network (CNN) and recurrent neural network (RNN), i.e., CNN+RNN, such as ResNet+LSTM \cite{he2016deep, hochreiter1997long}. 2) classic 3D CNNs, including 3D ResNet \cite{hara2018can}, C3D \cite{tran2015learning}, P3D \cite{qiu2017learning}, and R(2+1)D \cite{tran2018closer}. 3) hybrid architectures of CNN and Transformer, e.g., Former-DFER \cite{zhao2021former}, STT \cite{ma2022spatio}, NR-DFERNet \cite{li2022nr}, IAL \cite{li2023intensity}, and T-ESFL \cite{liu2022mafw}.

As shown in Table \ref{tab_dfew_sota}, we observe that SVFAP-B outperforms previous state-of-the-art methods on DFEW significantly (i.e., 5.72\% UAR and 5.02\% WAR improvement), setting a new record on this dataset.
Besides, the small version, SVFAP-S, also surpasses the best-performing methods by a large margin, achieving a better accuracy-complexity trade-off. When comparing the fine-grained performance of each class, we find that our methods achieve remarkable improvements for most emotions (e.g., \textit{happy} and \textit{sad}). Notably, for the rare \textit{disgust} emotion which only accounts for 1.2\% (about 146 samples) in the whole dataset, most baseline methods fail to classify its samples correctly. Nevertheless, our SVFAP-B improves the previous best performer by about 10\%. This result demonstrates that the proposed method can learn generic affect-related representations via large-scale self-supervised pre-training, thus alleviating the unbalanced learning in minority classes. 
Moreover, we have similar observations on the other two datasets. 
On the largest DFER dataset FERV39k, as shown in Table \ref{tab_ferv39k_sota}, SVFAP-B achieves 42.14\% UAR and 52.29\% WAR, outperforming the best baselines by 4.38\% UAR and 3.75\% WAR.
On the MAFW dataset, as given in Table \ref{tab_mafw_sota}, SVFAP-B improves over the state-of-the-art T-ESFL by 7.91\% UAR and 6.10\% WAR. 
Besides, a slight performance drop is also observed for both datasets.
To sum up, the above encouraging results on three in-the-wild datasets verify the strong generalization ability of our SVFAP in real-world scenarios. 

Finally, we present the comparison results on three small lab-controlled datasets, including CREMA-D, RAVDESS, and eNTERFACE05. The results  are reported in Table \ref{tab_cremad_sota}, \ref{tab_ravdess_sota}, and \ref{tab_enterface_sota}, respectively.
Similarly, we observe consistently significant performance improvements on these datasets. For instance, as shown in Table \ref{tab_cremad_sota}, SVFAP-B outperforms the best-performing unimodal methods by about 12\% UAR and 10\% WAR on CREMA-D. We also report the results of several multimodal methods. Compared with them, SVFAP-B still shows slightly better performance without using the audio information, which demonstrates the overwhelming superiority of the proposed method again.

\begin{table}[]
\caption{Comparison with state-of-the-art methods on CREMA-D.}
\label{tab_cremad_sota}
\centering
% \resizebox{\linewidth}{!}{
\begin{tabular}{lccc}
\toprule
Method           &   Modality         & UAR             & WAR            \\ \midrule
VO-LSTM \cite{ghaleb2019multimodal}           &   Video          & -                       & 66.80                    \\
Goncalves et al. \cite{goncalves2022robust}   &   Video          & -                       & 62.20                    \\
Lei et al. \cite{lei2023audio}                &   Video          & 64.68                   & 64.76                    \\
AV-LSTM \cite{ghaleb2019multimodal}           &   Video+Audio    & -                       & 72.90                    \\  
AV-Gating \cite{ghaleb2019multimodal}         &   Video+Audio    & -                       & 74.00                    \\
TFN \cite{zadeh2017tensor}                    &   Video+Audio    & -                       & 63.09                    \\
EF-GRU \cite{tran2022pre}                     &   Video+Audio    & -                       & 57.06                    \\
LF-GRU \cite{tran2022pre}                     &   Video+Audio    & -                       & 58.53                    \\
MulT Base \cite{tran2022pre}                  &   Video+Audio    & -                       & 68.87                    \\ 
MulT Large \cite{tran2022pre}                 &   Video+Audio    & -                       & 70.22                    \\ 
Goncalves et al. \cite{goncalves2022robust}   &   Video+Audio    & -                       & 77.30                    \\
\midrule
SVFAP-S (ours)                                &   Video          & 74.58                   & 74.58                    \\ 
SVFAP-B (ours)                                &   Video          & \textbf{77.31}          & \textbf{77.37}                    \\ 
\bottomrule
\end{tabular}
%}
\end{table}

\begin{table}[]
\caption{Comparison with state-of-the-art methods on RAVDESS.}
\label{tab_ravdess_sota}
\centering
% \resizebox{\linewidth}{!}{
\begin{tabular}{lccc}
\toprule
Method           &   Modality         & UAR             & WAR            \\ \midrule
VO-LSTM \cite{ghaleb2019multimodal}           &   Video          & -                       & 60.50                    \\
3D ResNeXt-50 \cite{su2020msaf}                &   Video          & -                       & 62.99                    \\
AV-LSTM \cite{ghaleb2019multimodal}           &   Video+Audio    & -                       & 65.80                    \\
AV-Gating \cite{ghaleb2019multimodal}         &   Video+Audio    & -                       & 67.70                    \\
MCBP \cite{su2020msaf}                        &   Video+Audio    & -                       & 71.32                   \\
MMTM \cite{su2020msaf}                        &   Video+Audio    & -                       & 73.12                   \\
MSAF \cite{su2020msaf}                        &   Video+Audio    & -                       & 74.86                   \\
CFN-SR \cite{fu2021cross}                     &   Video+Audio    & -                       & 75.76                   \\
\midrule
SVFAP-S (ours)                                &   Video          & 73.59                   & 73.90                    \\ 
SVFAP-B (ours)                                &   Video          & \textbf{75.15}                   & \textbf{75.01}                     \\ 
\bottomrule
\end{tabular}
%}
\end{table}

\begin{table}[]
\caption{Comparison with state-of-the-art methods on eNTERFACE05.}
\label{tab_enterface_sota}
\centering
% \resizebox{\linewidth}{!}{
\begin{tabular}{lccc}
\toprule
Method                  & UAR             & WAR            \\ \midrule
Mansoorizadeh et al. \cite{mansoorizadeh2010multimodal}      & -                       & 37.00                    \\
3DCNN \cite{byeon2014facial}                                 & -                       & 41.05                    \\
3DCNN-DAP \cite{byeon2014facial}                             & -                       & 41.36                    \\
Zhalehpour et al. \cite{zhalehpour2016baum}                  & -                       & 42.16                    \\
FAN \cite{meng2019frame}                                     & -                       & 51.44                    \\
STA-FER \cite{pan2019adeep}                                  & -                       & 42.98                    \\
TSA-FER \cite{pan2019deep}                                   & -                       & 43.72                    \\
C-LSTM \cite{miyoshi2019facial}                              & -                       & 45.29                    \\
EC-LSTM \cite{miyoshi2021enhanced}                           & -                       & 49.26                    \\
Graph-Tran \cite{zhao2022spatial}                            & -                       & 54.62                    \\
\midrule
SVFAP-S (ours)                                               & 57.16                   & 57.12                    \\ 
SVFAP-B (ours)                                               & \textbf{60.58}                   & \textbf{60.54}                    \\ 
\bottomrule
\end{tabular}
%}
\end{table}

\subsection{Dimensional Emotion Recognition}

\subsubsection{Datasets}
\textbf{Werewolf-XL} \cite{zhang2021werewolf} is a spontaneous audio-visual dataset with a total of 890 minutes of videos recorded during competitive group interactions in Werewolf games. It contains 131,688 video clips from 129 subjects in 30 game sessions, including 14,632 samples from active speakers and the rest from listeners in the game. Werewolf-XL provides both self-reported categorical emotion labels and externally assessed dimensional emotion scores. In this paper, we only use 14,632 speaker samples and dimensional annotations. As the goal is to predict continuous scores, we formulate it as a regression problem as stated in Section \ref{sec_finetuning}. Besides, we adopt a 5-fold session-independent cross-validation protocol for evaluation.
Two types of standard metrics, i.e., Concordance Correlation Coefficient (CCC) and Pearson Correlation Coefficient (PCC), are reported in this paper. They are computed as follows:
\begin{equation}
\begin{split}
\textrm{PCC} &= \frac{\textrm{cov}(\mathbf{y}, \hat{\mathbf{y}})}{\sigma_{\hat{y}} \sigma_y} \\
\textrm{CCC} &= \frac{2 \textrm{cov}(\mathbf{y}, \hat{\mathbf{y}})}{\sigma_{\hat{y}}^2 + \sigma_y^2 + (\mu_{\hat{y}} - \mu_y)^2} \\
\end{split}
\label{eq_metric_ccc_pcc}
\end{equation}
where $\mu_{\hat{y}}$ and $\mu_y$ are mean values of the overall predictions $\hat{\mathbf{y}}$ and overall labels $\mathbf{y}$ respectively, $\sigma_{\hat{y}}$ and $\sigma_y$ are their standard deviations, and $\textrm{cov}(\mathbf{y}, \hat{\mathbf{y}})$ calculates the covariance between predictions and labels. 

\textbf{AVCAffe} \cite{sarkar2022avcaffe} is currently the largest audio-visual dataset with both affect and cognitive load attributes. It is recorded by simulating a remote work setting and contains more than 108 hours of video from 106 subjects. Each subject is asked to participate in 7 specially designed tasks. After each task, self-reported affect (i.e., arousal and valence) and cognitive load scores are collected as ground truth labels. We only predict arousal and valence scores on a scale of 0-4 and formulate it as a classification problem according to the original paper. Since the duration of each task is too long (7.5 minutes), each video has been segmented into multiple 6-second short clips for model training. We follow the paper to obtain video-level predictions by averaging clip-level scores and employ the weighted F1 score as the evaluation metric. The dataset provides an official split: 86 subjects for training and 20 subjects for test.

\begin{table}[]
\caption{Comparison with state-of-the-art methods on Werewolf-XL.}
\label{tab_werewolfxl_sota}
\centering
\resizebox{\linewidth}{!}{
\begin{tabular}{lcccc}
\toprule
Method                                                    &  Modality                & Dimension     & PCC      & CCC            \\ \midrule
\multirow{3}{*}{HOG \cite{dalal2005histograms}}            &  \multirow{3}{*}{Video}  & Arousal       & 0.2082   & 0.1443         \\
                                                           &                          & Valence       & 0.5254   & 0.3456         \\
                                                           &                          & Dominance     & 0.2476   & 0.1690         \\ 
\multirow{3}{*}{VGGFace-LSTM \cite{zhang2021werewolf}}     &  \multirow{3}{*}{Video}  & Arousal       & 0.0724   & 0.0461         \\
                                                           &                          & Valence       & 0.6296   & 0.6038         \\
                                                           &                           & Dominance     & 0.1430   & 0.0820         \\ 
\multirow{3}{*}{Zhang et al. \cite{zhang2021werewolf}}     &  \multirow{3}{*}{Video+Audio}  & Arousal       & 0.1641   & 0.2770         \\
                                                           &                          & Valence       & 0.6314   & 0.6234         \\
                                                           &                          & Dominance     & 0.3540   & 0.3840         \\
\midrule
\multirow{3}{*}{SVFAP-S (ours)}                            &  \multirow{3}{*}{Video}  & Arousal       & 0.2211   & 0.1786         \\
                                                           &                          & Valence       & 0.6566   & 0.6374         \\
                                                           &                          & Dominance     & 0.3202   & 0.2799         \\ 
\multirow{3}{*}{SVFAP-B (ours)}                            &  \multirow{3}{*}{Video}  & Arousal      & \textbf{0.2351}  & \textbf{0.1896}   \\
                                                           &                          & Valence      & \textbf{0.6711}  & \textbf{0.6427}    \\
                                                           &                          & Dominance    & \textbf{0.3461}  & \textbf{0.2969}    \\
\bottomrule
\end{tabular}
}
\end{table}

\begin{table}[]
\caption{Comparison with state-of-the-art methods on AVCAffe.}
\label{tab_avcaffe_sota}
\centering
\resizebox{\linewidth}{!}{
\begin{tabular}{lccc}
\toprule
Method           &   Modality         & Arousal             & Valence            \\ \midrule
MC3-18 \cite{tran2018closer}                           &   Video          & 34.00                   & 38.80                    \\
3D ResNet-18 \cite{hara2018can}                        &   Video          & 30.90                   & 39.50                    \\
R(2+1)D-18 \cite{tran2018closer}                       &   Video          & 33.30                   & 34.90                    \\
MC3-18+VGG-16 \cite{sarkar2022avcaffe}                 &   Video+Audio    & 38.90                   & 41.70                    \\
3D ResNet-18+VGG-16 \cite{sarkar2022avcaffe}           &   Video+Audio    & 37.30                   & 39.40                    \\
R(2+1)D-18+VGG-16 \cite{sarkar2022avcaffe}             &   Video+Audio    & 40.50                   & 39.50                    \\
MC3-18+ResNet-18 \cite{sarkar2022avcaffe}              &   Video+Audio    & 36.00                   & 39.20                    \\
3D ResNet-18+ResNet-18 \cite{sarkar2022avcaffe}        &   Video+Audio    & 35.10                   & 39.10                    \\
R(2+1)D-18+ResNet-18 \cite{sarkar2022avcaffe}          &   Video+Audio    & 39.50                   & 37.70                    \\
\midrule
SVFAP-S (ours)                                &   Video          & 39.24                   & 40.22                    \\ 
SVFAP-B (ours)                                &   Video          & \textbf{40.36}                   & \textbf{41.49}               \\ 
\bottomrule
\end{tabular}
}
\end{table}

\subsubsection{Comparison with State-of-the-art Methods}
The results on Werewolf-XL are shown in Table \ref{tab_werewolfxl_sota}. As we can see, our method achieves much better performance than unimodal baseline methods in three emotion dimensions. 
Specifically, SVFAP-B outperforms the best-performing ones by about 0.04 CCC and 0.03 PCC in arousal, 0.04 CCC and 0.04 PCC in valence, and 0.13 CCC and 0.10 PCC in dominance. 
Besides, we also show the multimodal baseline in the table. Compared with it, SVFAP-B still shows superior performance in valence, although worse results are achieved in arousal and dominance. Besides, we only observe a moderate performance degradation for SVFAP-S on this dataset.

We further present the comparison results on AVCAffe in Table \ref{tab_avcaffe_sota}. Similarly, we observe that both SVFAP-B and SVFAP-S surpass the state-of-the-art unimodal methods by a large margin (especially in arousal). For instance, SVFAP-B achieves about 6\% F1-score improvement in arousal and about 2\% in valence.
Moreover, when compared with multimodal methods, our base model still presents a comparable performance in both arousal and valence, which verifies the effectiveness of large-scale self-supervised pre-training for dimension emotion recognition.

\subsection{Personality Recognition}

\begin{table}[]
\caption{Comparison with state-of-the-art methods on ChaLearn 2016 First Impression in terms of CCC. O: Openness. C: Conscientiousness. E: Extraversion. A: Agreeableness. N: Neuroticism.}
\label{tab_chalearn_sota_ccc}
\centering
\resizebox{\linewidth}{!}{
\begin{tabular}{lcccccc}
\toprule
Method                  & O   & C  & E & A  & N  & Average            \\ \midrule
DAN \cite{wei2017deep}                        & 0.5693 & 0.6254 & 0.6070 & 0.4855 & 0.6025 & 0.5779               \\
ResNet \cite{guccluturk2016deep}              & 0.1561 & 0.1902 & 0.1355 & 0.0838 & 0.1373 & 0.1406  \\
CRNet \cite{li2020cr}                         & 0.3748 & 0.3646 & 0.3987 & 0.2390 & 0.3226 & 0.3399  \\
CAM-DAN$_+$ \cite{ventura2017interpreting}    & 0.5882 & 0.6550 & 0.6326 & 0.5003 & 0.6199 & 0.5992  \\
PersEmoN \cite{zhang2019persemon}             & 0.2067 & 0.2441 & 0.2675 & 0.1369 & 0.1768 & 0.2064  \\
Amb-Fac \cite{suman2022multi}                 & 0.5858 & 0.6750 & 0.5997 & 0.4971 & 0.5765 & 0.5868  \\
SENet \cite{hu2018squeeze}                    & 0.5300 & 0.5580 & 0.5815 & 0.4493 & 0.5708 & 0.5379  \\
HRNet \cite{wang2020deep}                     & 0.5923 & 0.6912 & 0.6436 & 0.5195 & 0.6273 & 0.6148  \\
Swin \cite{liu2021swin}                       & 0.2223 & 0.2426 & 0.2531 & 0.1224 & 0.1942 & 0.2069  \\
3D ResNet \cite{hara2018can}                  & 0.3248 & 0.3601 & 0.3601 & 0.2120 & 0.3352 & 0.3185  \\
Slow-Fast \cite{feichtenhofer2019slowfast}    & 0.0256 & 0.0320 & 0.0185 & 0.0105 & 0.0184 & 0.0210  \\
TPN \cite{yang2020temporal}                   & 0.4427 & 0.4767 & 0.4998 & 0.3230 & 0.4675 & 0.4420  \\
VAT \cite{girdhar2019video}                   & 0.6216 & 0.6753 & 0.6836 & 0.5228 & 0.6456 & 0.6298  \\
\midrule
SVFAP-S (ours)                                & 0.6313 & 0.6974 & 0.7210 & 0.5427 & 0.6648 & 0.6514  \\
SVFAP-B (ours)                                & \textbf{0.6511} & \textbf{0.7141} & \textbf{0.7351} & \textbf{0.5498} & \textbf{0.6724} & \textbf{0.6645}  \\
\bottomrule
\end{tabular}
}
\end{table}

\begin{table}[]
\caption{Comparison with state-of-the-art methods on ChaLearn 2016 First Impression in terms of ACC. O: Openness. C: Conscientiousness. E: Extraversion. A: Agreeableness. N: Neuroticism.}
\label{tab_chalearn_sota_acc}
\centering
\resizebox{\linewidth}{!}{
\begin{tabular}{lcccccc}
\toprule
Method                  & O   & C  & E & A  & N  & Average            \\ \midrule
DAN \cite{wei2017deep}                        & 0.9098 & 0.9106 & 0.9096 & 0.9102 & 0.9061 & 0.9093               \\
CNN-LSTM \cite{subramaniam2016bi}             & 0.8832 & 0.8742 & 0.8778 & 0.8933 & 0.8770 & 0.8811               \\
ResNet \cite{guccluturk2016deep}              & 0.8896 & 0.8835 & 0.8837 & 0.8968 & 0.8830 & 0.8873  \\
CRNet \cite{li2020cr}                         & 0.8987 & 0.8932 & 0.8952 & 0.9018 & 0.8908 & 0.8960  \\
CAM-DAN$_+$ \cite{ventura2017interpreting}    & 0.9115 & 0.9139 & 0.9126 & 0.9118 & 0.9089 & 0.9118  \\
PersEmoN \cite{zhang2019persemon}             & 0.8934 & 0.8893 & 0.8913 & 0.8994 & 0.8866 & 0.8920  \\
Amb-Fac \cite{suman2022multi}                 & 0.9101 & 0.9141 & 0.9082 & 0.9095 & 0.9038 & 0.9091  \\
SENet \cite{hu2018squeeze}                    & 0.9076 & 0.9060 & 0.9080 & 0.9097 & 0.9061 & 0.9075  \\
HRNet \cite{wang2020deep}                     & 0.9101 & 0.9154 & 0.9111 & 0.9113 & 0.9084 & 0.9113  \\
Swin \cite{liu2021swin}                       & 0.8937 & 0.8870 & 0.8893 & 0.8983 & 0.8860 & 0.8909  \\
3D ResNet \cite{hara2018can}                  & 0.8964 & 0.8921 & 0.8933 & 0.9008 & 0.8915 & 0.8948  \\
Slow-Fast \cite{feichtenhofer2019slowfast}    & 0.8780 & 0.8604 & 0.8443 & 0.8809 & 0.8613 & 0.8650  \\
TPN \cite{yang2020temporal}                   & 0.9025 & 0.8963 & 0.9019 & 0.9013 & 0.8992 & 0.9003  \\
VAT \cite{girdhar2019video}                   & 0.9115 & 0.9123 & 0.9153 & 0.9099 & 0.9098 & 0.9118  \\
\midrule
SVFAP-S (ours)                                & 0.9144 & 0.9175 & 0.9208 & 0.9135 & 0.9134 & 0.9159  \\
SVFAP-B (ours)                                & \textbf{0.9162} & \textbf{0.9187} & \textbf{0.9227} & \textbf{0.9152} & \textbf{0.9145} & \textbf{0.9175}  \\
\bottomrule
\end{tabular}
}
\end{table}

Finally, to further verify the general applicability of the proposed method, we evaluate it on the personality recognition task. Different from the above two tasks we have explored, this task requires stable and prototypical facial behavior modeling to capture relevant features that reflect personality traits. Thus, experiments on this task can provide a more comprehensive evaluation of our proposed method. 

The classic ChaLearn First Impression dataset \cite{ponce2016chalearn} is selected for evaluation. It consists of 10,000 talking-to-the-camera clips extracted from over 3,000 high-definition YouTube videos. Each video clip has a duration of about 15 seconds and is annotated with the Big Five personality traits (i.e., openness, conscientiousness, extraversion, agreeableness, and neuroticism). This dataset has been split into three sets: 6000 videos for training, 2000 videos for validation, and 2000 videos for the final test. Since the task is to predict continuous scores in different personality dimensions, we formulate it as a regression problem. Following \cite{liao2024open}, we use two standard evaluation metrics, i.e., CCC and the Accuracy (ACC). CCC is defined in Eq. (\ref{eq_metric_ccc_pcc}). The definition of ACC is given as follows:
\begin{equation}
\begin{split}
\textrm{ACC} = 1 - \frac{1}{N} \sum_{i=1}^{N}|y_i - \hat{y}_i|\\
\end{split}
\label{eq_metric_acc}
\end{equation}
where $\hat{y}$ is the prediction, $y$ is the label, and $N$ is the number of test videos.

The comparison results in terms of CCC and ACC are shown in Table \ref{tab_chalearn_sota_ccc} and Table \ref{tab_chalearn_sota_acc}, respectively. For CCC, we observe that two versions of our models achieve similar performance. Both of them outperform the best-performing supervised baseline largely, achieving about 0.02-0.03 CCC improvement. For ACC, we find that the variation between different methods is relatively small. Nevertheless, our methods still show the best performance.

\section{Conclusion}
In this paper, we have presented a self-supervised learning method, termed Self-supervised Video Facial Affect Perceiver (SVFAP), to unleash the power of large-scale self-supervised pre-training for video-based facial affect analysis. 
SVFAP utilizes masked facial video autoencoding as the objective to perform self-supervised pre-training on a large amount of unlabeled facial videos. Besides, it employs a novel TBSBT model as the encoder to minimize large redundancy in 3D facial video data from both spatial and temporal perspectives, leading to significantly lower computational costs and superior performance.
To verify the effectiveness of SVFAP, we conduct extensive experiments on nine datasets in three popular downstream tasks, including dynamic facial expression recognition, dimensional emotion recognition, and personality recognition.
The results demonstrate that SVFAP can learn powerful affect-related representations via large-scale self-supervised pre-training. Specifically, it largely outperforms previous pre-trained supervised and self-supervised models and also shows strong adaptation ability in the low data regime. Moreover, our SVFAP achieves significant improvements over state-of-the-art methods in three downstream tasks, setting new records on all datasets. 

In future work, we plan to investigate the scaling behavior of SVFAP using larger models and more unlabeled data. Besides, it is also interesting to evaluate SVFAP in other downstream tasks, such as dynamic micro-expression recognition and depression level detection. We also hope our work can inspire more relevant research to further advance the development of video-based facial affect analysis.

% if have a single appendix:
%\appendix[Proof of the Zonklar Equations]
% or
%\appendix  % for no appendix heading
% do not use \section anymore after \appendix, only \section*
% is possibly needed

% use appendices with more than one appendix
% then use \section to start each appendix
% you must declare a \section before using any
% \subsection or using \label (\appendices by itself
% starts a section numbered zero.)
%

% \appendices
% \section{Proof of the First Zonklar Equation}
% Appendix one text goes here.

% % you can choose not to have a title for an appendix
% % if you want by leaving the argument blank
% \section{}
% Appendix two text goes here.

% use section* for acknowledgment
\ifCLASSOPTIONcompsoc
  % The Computer Society usually uses the plural form
  \section*{Acknowledgments}
\else
  % regular IEEE prefers the singular form
  \section*{Acknowledgment}
\fi

% The authors would like to thank...

% This work was supported in part by the National Natural Science Foundation of China (NSFC) under Grant 61831022, Grant U21B2010, Grant 61901473, and Grant 62101553; and in part by the Open Research Projects of Zhejiang Laboratory under Grant 2021KH0AB06.

% This work was supported in part by the National Natural Science Foundation of China (NSFC) (No.61831022, No.U21B2010, No.61901473, No.62101553) and the Open Research Projects of Zhejiang Lab (NO. 2021KH0AB06).

% This work is supported by the National Natural Science Foundation of China (NSFC) (No.62201572, No.61831022, No.62276259, No.U21B2010), Beijing Municipal Science \& Technology Commission, Administrative Commission of Zhongguancun Science Park (No.Z211100004821013), Open Research Projects of Zhejiang Lab (No.2021KH0AB06), CCF-Baidu Open Fund (No.OF2022025).

This work is supported by the National Natural Science Foundation of China (NSFC) ( No.62276259, No.62201572, No.U21B2010, No.62271083, No.62306316).

% Can use something like this to put references on a page
% by themselves when using endfloat and the captionsoff option.
\ifCLASSOPTIONcaptionsoff
  \newpage
\fi

% trigger a \newpage just before the given reference
% number - used to balance the columns on the last page
% adjust value as needed - may need to be readjusted if
% the document is modified later
%\IEEEtriggeratref{8}
% The "triggered" command can be changed if desired:
%\IEEEtriggercmd{\enlargethispage{-5in}}

% references section

% can use a bibliography generated by BibTeX as a .bbl file
% BibTeX documentation can be easily obtained at:
% http://mirror.ctan.org/biblio/bibtex/contrib/doc/
% The IEEEtran BibTeX style support page is at:
% http://www.michaelshell.org/tex/ieeetran/bibtex/
\bibliographystyle{IEEEtran}
% argument is your BibTeX string definitions and bibliography database(s)
\bibliography{main}
%
% <OR> manually copy in the resultant .bbl file
% set second argument of \begin to the number of references
% (used to reserve space for the reference number labels box)
% \begin{thebibliography}{1}

% \bibitem{IEEEhowto:kopka}
% H.~Kopka and P.~W. Daly, \emph{A Guide to \LaTeX}, 3rd~ed.\hskip 1em plus
%   0.5em minus 0.4em\relax Harlow, England: Addison-Wesley, 1999.

% \end{thebibliography}

% biography section
% 
% If you have an EPS/PDF photo (graphicx package needed) extra braces are
% needed around the contents of the optional argument to biography to prevent
% the LaTeX parser from getting confused when it sees the complicated
% \includegraphics command within an optional argument. (You could create
% your own custom macro containing the \includegraphics command to make things
% simpler here.)
%\begin{IEEEbiography}[{\includegraphics[width=1in,height=1.25in,clip,keepaspectratio]{mshell}}]{Michael Shell}
% or if you just want to reserve a space for a photo:

\begin{IEEEbiography}[{\includegraphics[width=1.1in,height=1.25in,clip,keepaspectratio]{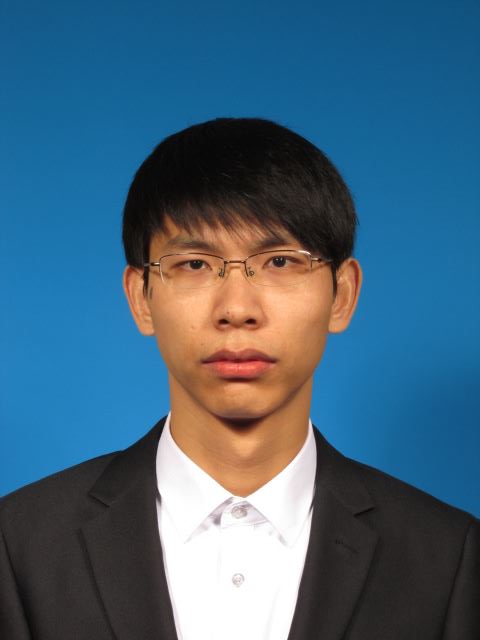}}]{Licai Sun}
received the B.S. degree from Beijing Forestry University, Beijing, China, in 2016, and the M.S. degree from University of Chinese Academy of Sciences, Beijing, China, in 2019. He is currently working toward the Ph.D. degree with the School of Artificial Intelligence, University of Chinese Academy of Sciences, Beijing,
China. His current research interests include affective computing, deep learning, and multimodal representation learning.
\end{IEEEbiography}

% if you will not have a photo at all:
\begin{IEEEbiography}[{\includegraphics[width=1.1in,height=1.25in,clip,keepaspectratio]{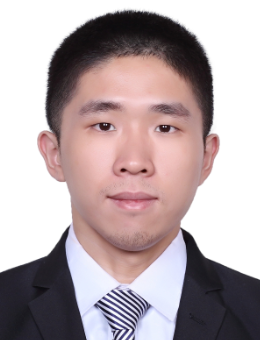}}]{Zheng Lian}
received the B.S. degree from the Beijing University of Posts and Telecommunications (BUPT), Beijing, China, in 2016. And he received the Ph.D. degree from the Institute of Automation, Chinese Academy of Sciences, Beijing, China, in 2021. He is currently an Assistant Professor at National Laboratory of Pattern Recognition, Institute of Automation, Chinese Academy of Sciences, Beijing, China. His current research interests include affective computing, deep learning, and multimodal emotion recognition.
\end{IEEEbiography}

\begin{IEEEbiography}[{\includegraphics[width=1.1in,height=1.25in,clip,keepaspectratio]{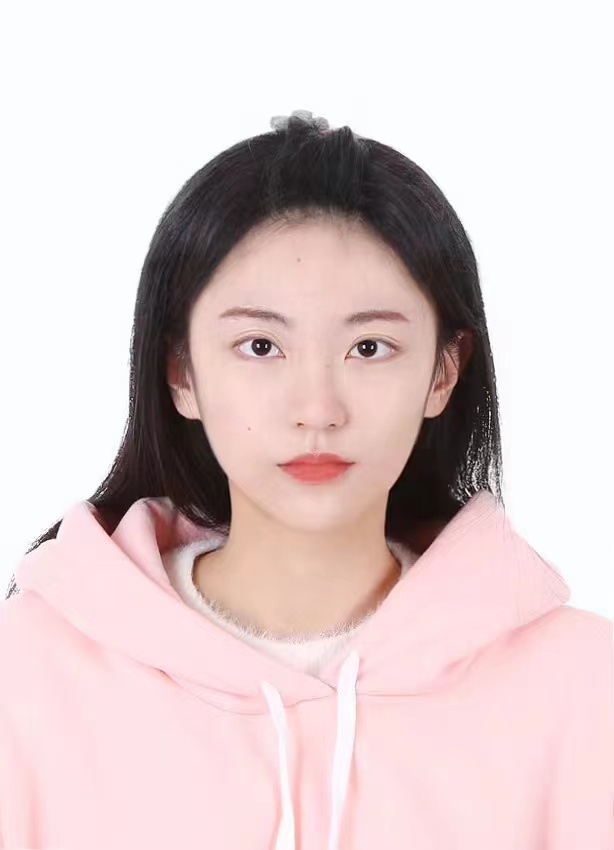}}]{Kexin Wang}
received the B.S. degree from Beijing University of Aeronautics and Astronautics, Beijing, China, in 2021. She is currently working toward the M.S. degree with the Institute of Automation, China Academy of Sciences, Beijing, China. Her current research interests include multi-label emotion recognition and noisy label learning.
\end{IEEEbiography}

\begin{IEEEbiography}[{\includegraphics[width=1.1in,height=1.25in,clip,keepaspectratio]{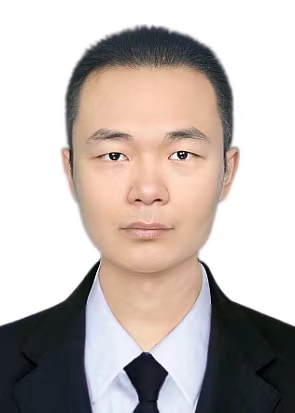}}]{Yu He}
received the B.S. degree from Hunan University, China, in 2013. He is currently working toward the M.S. degree with the University of Chinese Academy of Sciences, Beijing, China. His current research interests include multimodal affective computing and physiological signal prediction.
\end{IEEEbiography}

\begin{IEEEbiography}[{\includegraphics[width=1.1in,height=1.25in,clip,keepaspectratio]{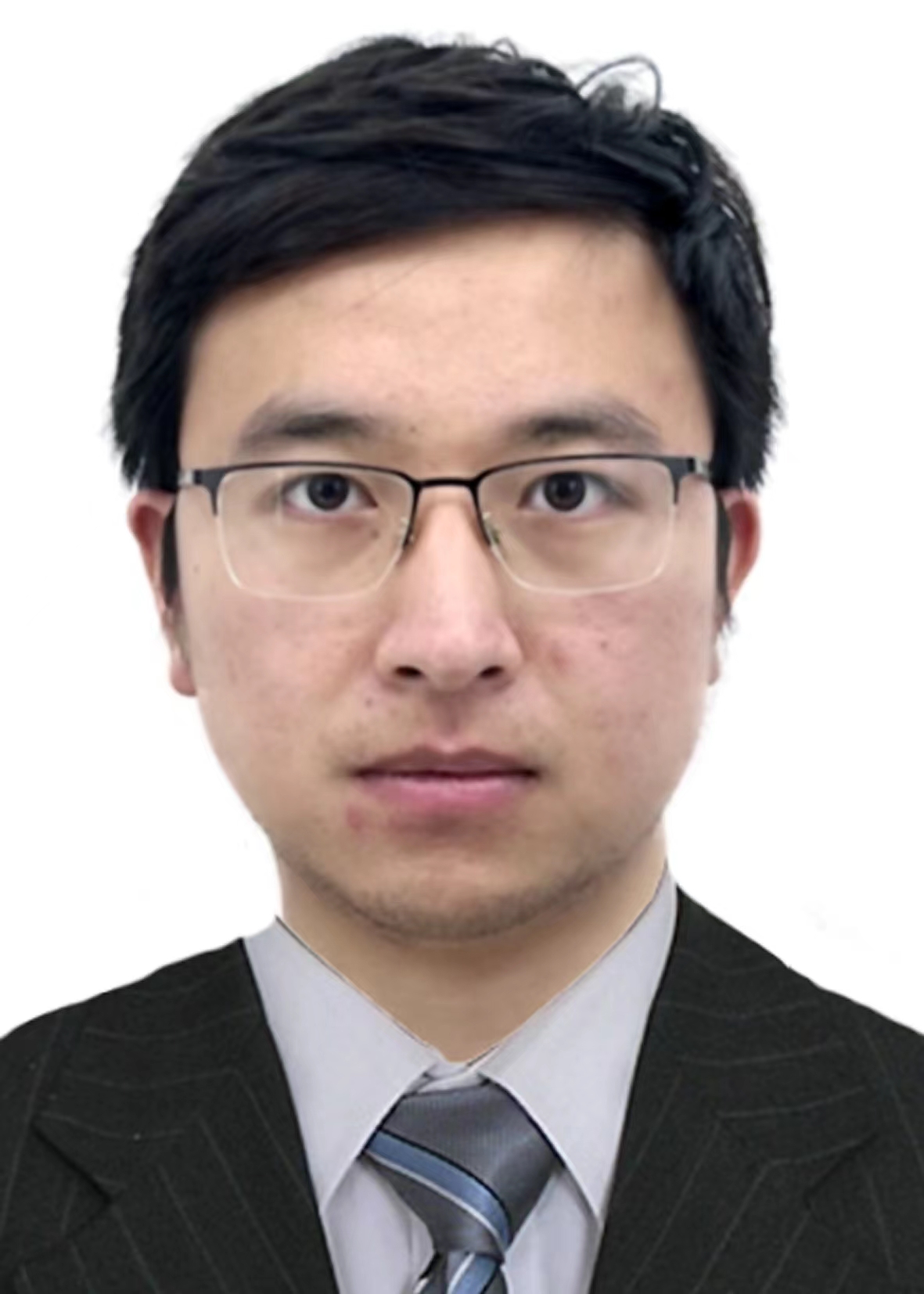}}]{Mingyu Xu}
received the B.S. degree from Peking University, Beijing, China, in 2021. He is currently working toward the M.S. degree with the Institute of Automation, China Academy of Sciences, Beijing, China. His current research interests include uncertainty learning and partial label learning.
\end{IEEEbiography}

\begin{IEEEbiography}[{\includegraphics[width=1.1in,height=1.25in,clip,keepaspectratio]{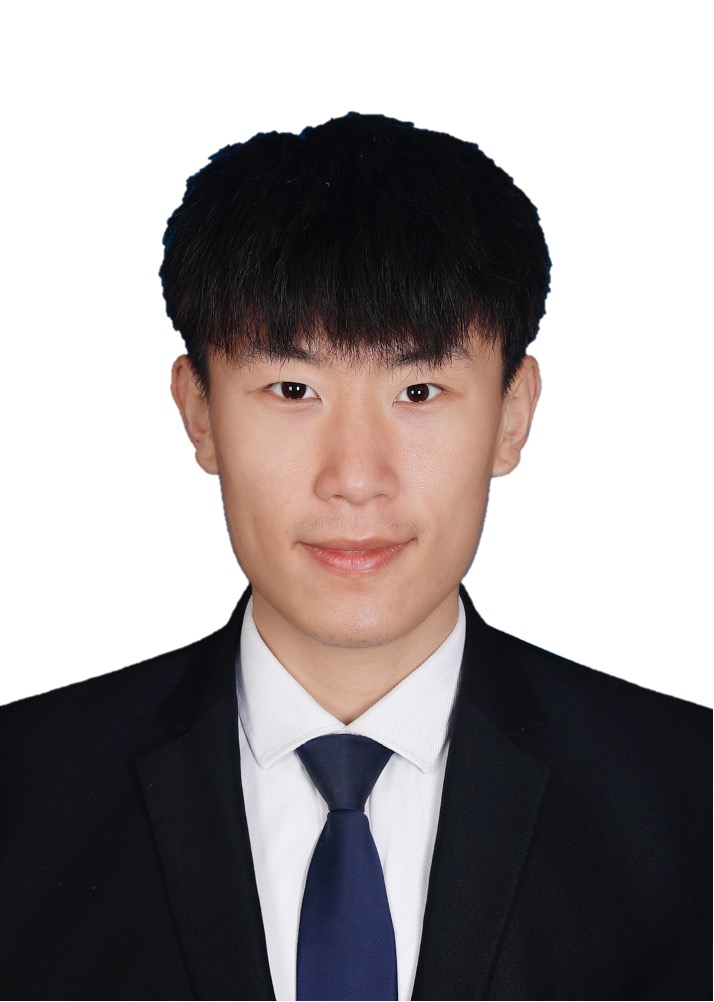}}]{Haiyang Sun}
received the B.E. degree from Shandong University of Science and Technology, China, in 2021. He is currently working toward the M.S. degree with the Institute of Automation, China Academy of Sciences, Beijing, China. His current research interests include multimodal emotion recognition and neural architecture search.
\end{IEEEbiography}

\begin{IEEEbiography}[{\includegraphics[width=1.1in,height=1.25in,clip,keepaspectratio]{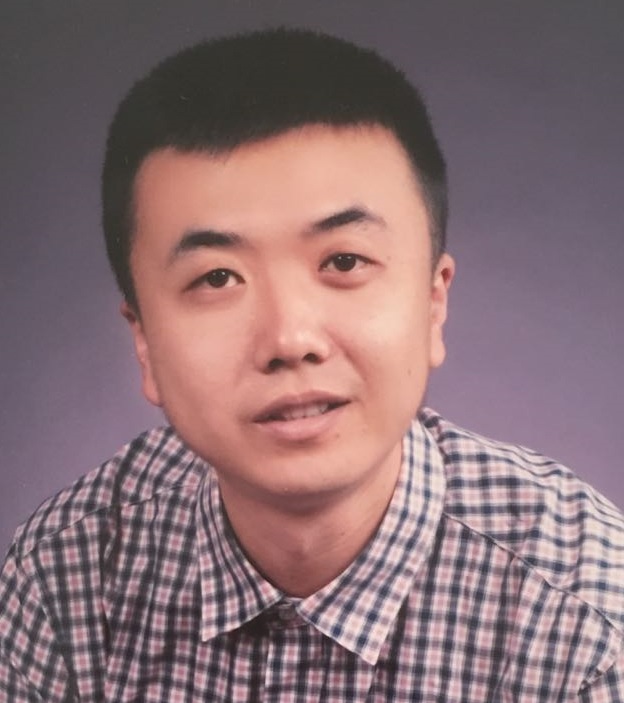}}]{Bin Liu}
	received his the B.S. degree and the M.S. degree from Beijing institute of technology (BIT), Beijing, China, in 2007 and 2009 respectively. He received Ph.D. degree from the National Laboratory of Pattern Recognition, Institute of Automation, Chinese Academy of Sciences, Beijing, China, in 2015. He is currently an Associate Professor in the National Laboratory of Pattern Recognition, Institute of Automation, Chinese Academy of Sciences, Beijing, China. His current research interests include affective computing and audio signal processing.
\end{IEEEbiography}

\begin{IEEEbiography}[{\includegraphics[width=1.1in,height=1.25in,clip,keepaspectratio]{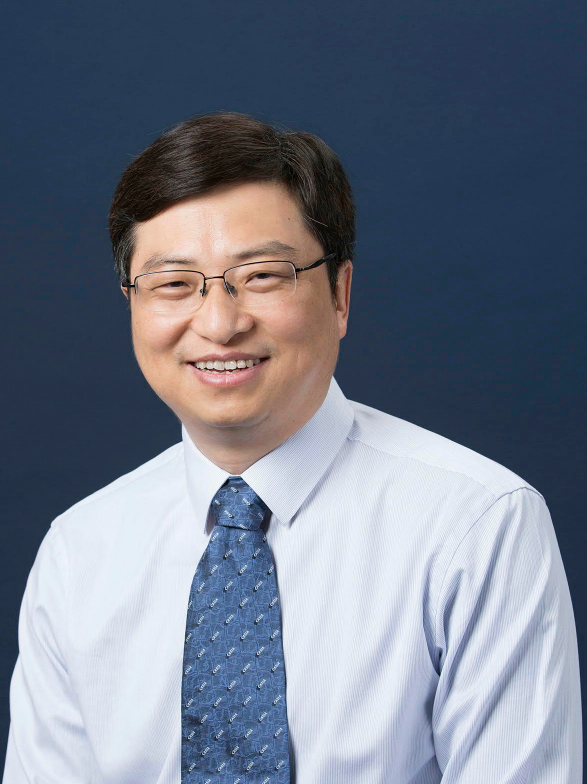}}]{Jianhua Tao}
	received the Ph.D. degree from Tsinghua University, Beijing, China, in 2001, and the M.S. degree from Nanjing University, Nanjing, China, in 1996. He is currently a Professor with Department of Automation, Tsinghua University, Beijing, China. He has authored or coauthored more than eighty papers on major journals and proceedings. His current research interests include speech recognition, speech synthesis and coding methods, human–computer interaction, multimedia information processing, and pattern recognition. He is the Chair or Program Committee Member for several major conferences, including ICPR, ACII, ICMI, ISCSLP, etc. He is also the Steering Committee Member for the IEEE Transactions on Affective Computing, an Associate Editor for Journal on Multimodal User Interface and International Journal on Synthetic Emotions, and the Deputy Editor-in-Chief for Chinese Journal of Phonetics.
\end{IEEEbiography}

\end{document}